\documentclass{article}
\usepackage{iclr2027_conference,times}
\usepackage{hyperref} 
\hypersetup{hidelinks}

\usepackage[table,dvipsnames]{xcolor}
\definecolor{lightgray}{gray}{0.90}

\usepackage{amsmath,amsfonts,bm}

\def\eqref#1{equation~\ref{#1}}

\def\1{\bm{1}}

\DeclareMathAlphabet{\mathsfit}{\encodingdefault}{\sfdefault}{m}{sl}
\SetMathAlphabet{\mathsfit}{bold}{\encodingdefault}{\sfdefault}{bx}{n}

\definecolor{component1}{HTML}{85200c}
\definecolor{component2}{HTML}{0b5394}

\newcommand{\componenta}{\textcolor{component2}}
\newcommand{\componentb}{\textcolor{component1}}
\usepackage{amssymb}
\usepackage{mathtools}
\usepackage{amsthm}
\usepackage{nicefrac}

\usepackage{microtype}
\usepackage{xspace}
\usepackage{pifont}
\usepackage[capitalize,noabbrev]{cleveref}

\usepackage{graphicx}
\usepackage{subcaption}
\usepackage{wrapfig}
\usepackage{float}
\usepackage{placeins}

\usepackage{booktabs}
\usepackage{multirow}
\usepackage{tabularx}

\usepackage[ruled]{algorithm2e}

\usepackage{url}
\usepackage[capitalize,noabbrev]{cleveref}

\title{Learning Not to Interfere: Adapting Experts for Model Merging}

\author{%
  \parbox{\dimexpr\textwidth-2\tabcolsep\relax}{%
    \centering
    \resizebox{\linewidth}{!}{%
      \normalfont\normalsize\bfseries
      Pratik Ramesh\textsuperscript{1}\thanks{Corresponding author:
        \href{mailto:pramesh39@gatech.edu}{ \textcolor{RoyalBlue}{\texttt{pramesh39@gatech.edu}}}.}\quad
      George Stoica\textsuperscript{1,2}\quad
      Arun Iyer\textsuperscript{3}\quad
      Leshem Choshen\textsuperscript{4,5}\quad
      Judy Hoffman\textsuperscript{1,6}%
    }\\[0.7em]
    \resizebox{\linewidth}{!}{%
      \normalfont\small
      \textsuperscript{1}Georgia Tech\quad
      \textsuperscript{2}University of Washington\quad
      \textsuperscript{3}Microsoft Research\quad
      \textsuperscript{4}MIT\quad
      \textsuperscript{5}IBM Research\quad
      \textsuperscript{6}University of California, Irvine%
    }%
    \endgraf
  }%
}
\date{}

\hypersetup{
  pdftitle={Learning Not to Interfere: Adapting Experts for Model Merging},
  pdfauthor={Pratik Ramesh, George Stoica, Arun Iyer, Leshem Choshen, Judy Hoffman}
}

\newif\ifshowcomments
\showcommentsfalse

\newcommand{\method}{\textsc{RI}\xspace}

\iclrfinalcopy

\begin{document}
\maketitle

\fancyhead{}
\fancyhead[L]{\small\textit{Preprint}}
\renewcommand{\headrulewidth}{0.4pt}

\newcommand{\FigTeaser}{%
\begin{wrapfigure}{r}{0.4\textwidth}
  \centering
  \vspace{-30pt}
  \includegraphics[width=\linewidth]{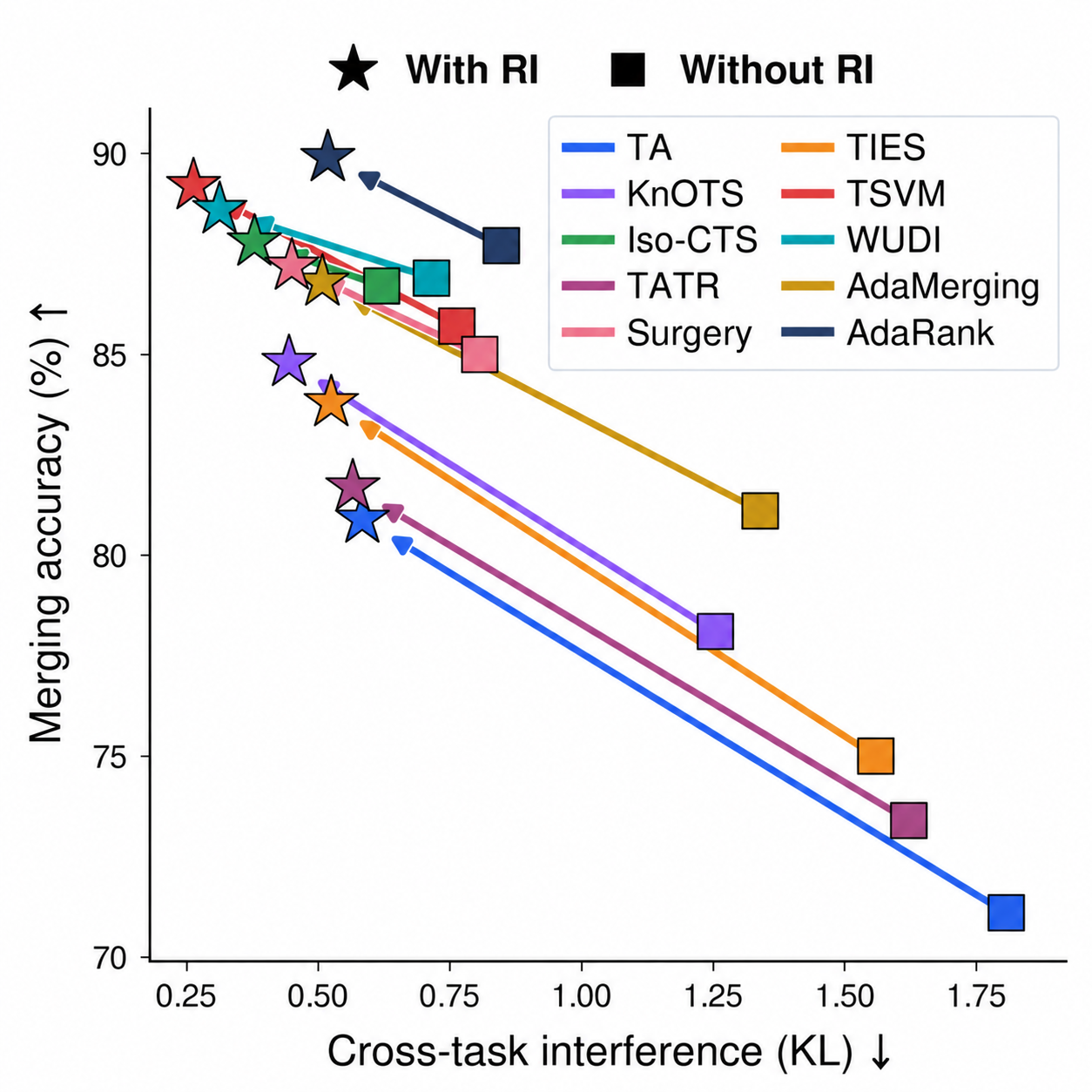}
  \caption{
    \textbf{Resolving Interference (\method)} is a lightweight adaptation strategy
    that mitigates cross-task interference, enhancing the performance of existing
    model-merging techniques.
  }
    \vspace{-10pt}
  \label{fig:teaser}
\end{wrapfigure}
}

\newcommand{\FigMethod}{%
\begin{figure*}[t]
    \centering
    \includegraphics[width=1.0\linewidth]{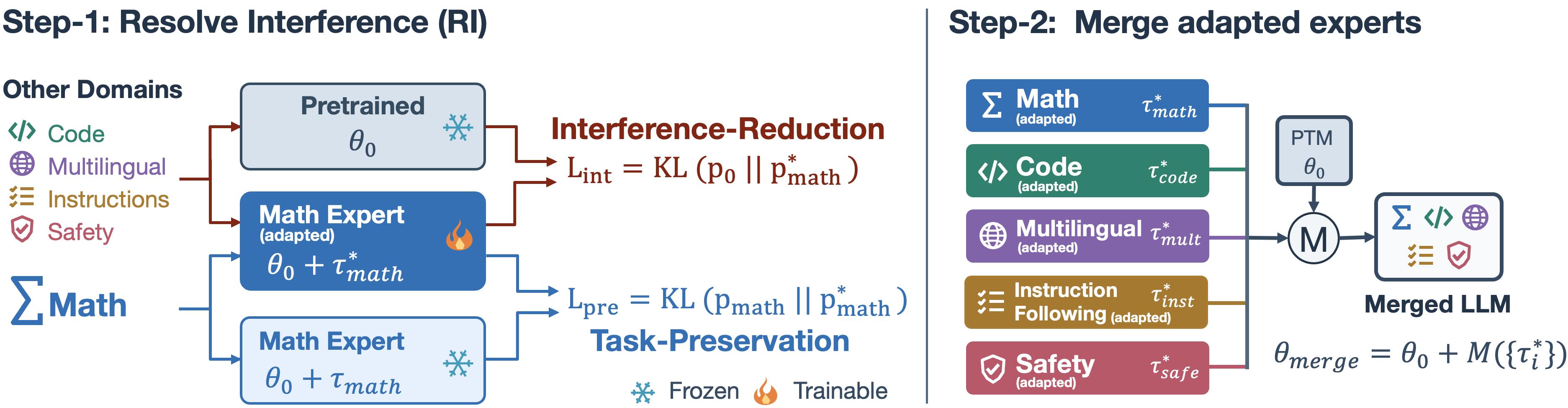}    
\caption{\textbf{Resolving Interference (RI) prepares experts for merging.}
\textbf{(1) Resolving Interference.} Illustrated for the math expert, RI uses unlabeled inputs and two KL distillation objectives: \emph{task preservation} matches the original expert's predictions on its own task, while \emph{interference reduction} matches the shared pretrained model's predictions on other tasks. Together, these objectives preserve specialization while restricting the task vector's influence on other tasks.
\textbf{(2) Merging adapted experts.} After adaptation any merging operator $M$ combines the adapted task vectors $\{\tau_i^\star\}$, which are added to the shared pretrained initialization $\theta_0$ to obtain a single multitask model.}
    \label{fig:method_figure}
\end{figure*}
}

\newcommand{\FigHeatmap}{%
\begin{figure*}[t]
    \centering
    \includegraphics[width=1.0\linewidth]{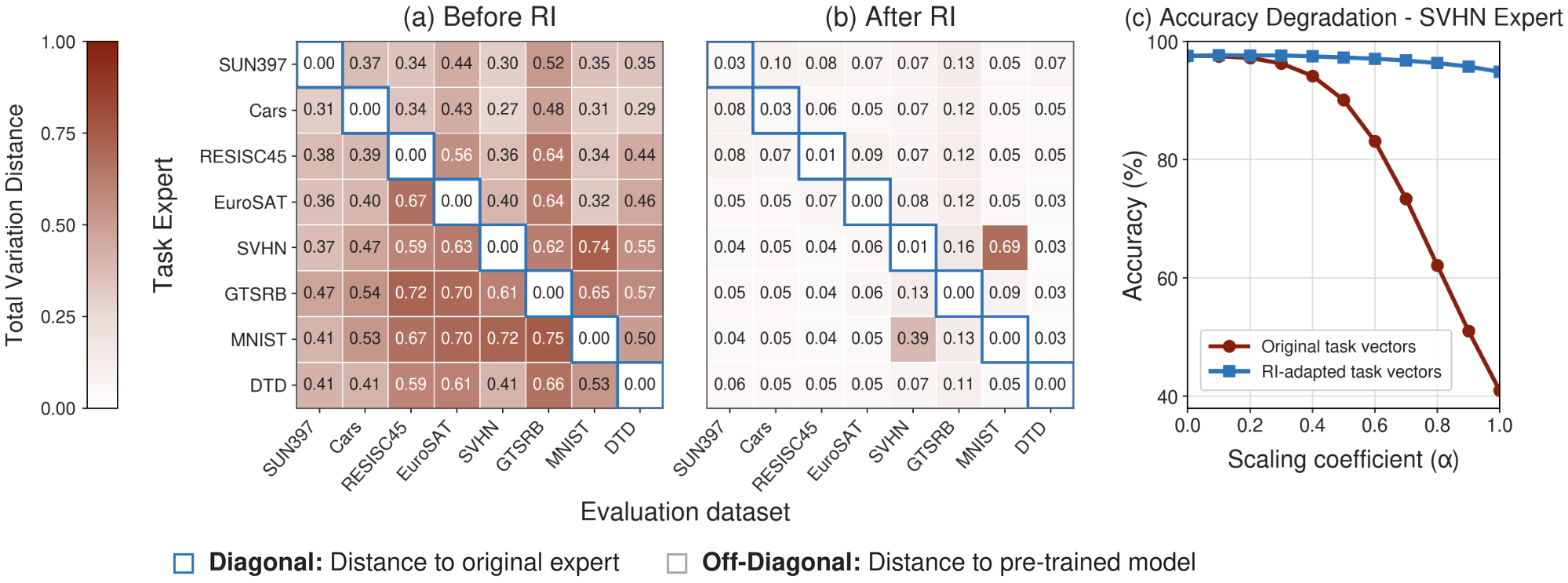}    

\caption{\textbf{RI preserves specialization and makes task vectors less disruptive when merged.}
\textbf{(a,b)} Heatmaps report mean total variation distances across experts (rows) and datasets (columns). Diagonal entries compare against the original expert; off-diagonal entries compare against the shared pretrained model, the reference before specialization. RI reduces changes on other tasks while preserving own-task predictions.
\textbf{(c)} Adding RI-adapted updates from other tasks causes substantially less accuracy degradation for the fixed, unscaled SVHN expert ($94.9\%$ versus $41.0\%$ with original updates at $\alpha=1$). Similar analysis for other experts models can be found in Figure~\ref{fig:tv_sweep_all}.}
    \vspace{-5pt}
    \label{fig:heatmap}
\end{figure*}
}

\newcommand{\FigConsistency}{%
\begin{figure*}[t]
    \centering
    \includegraphics[width=0.9\linewidth]{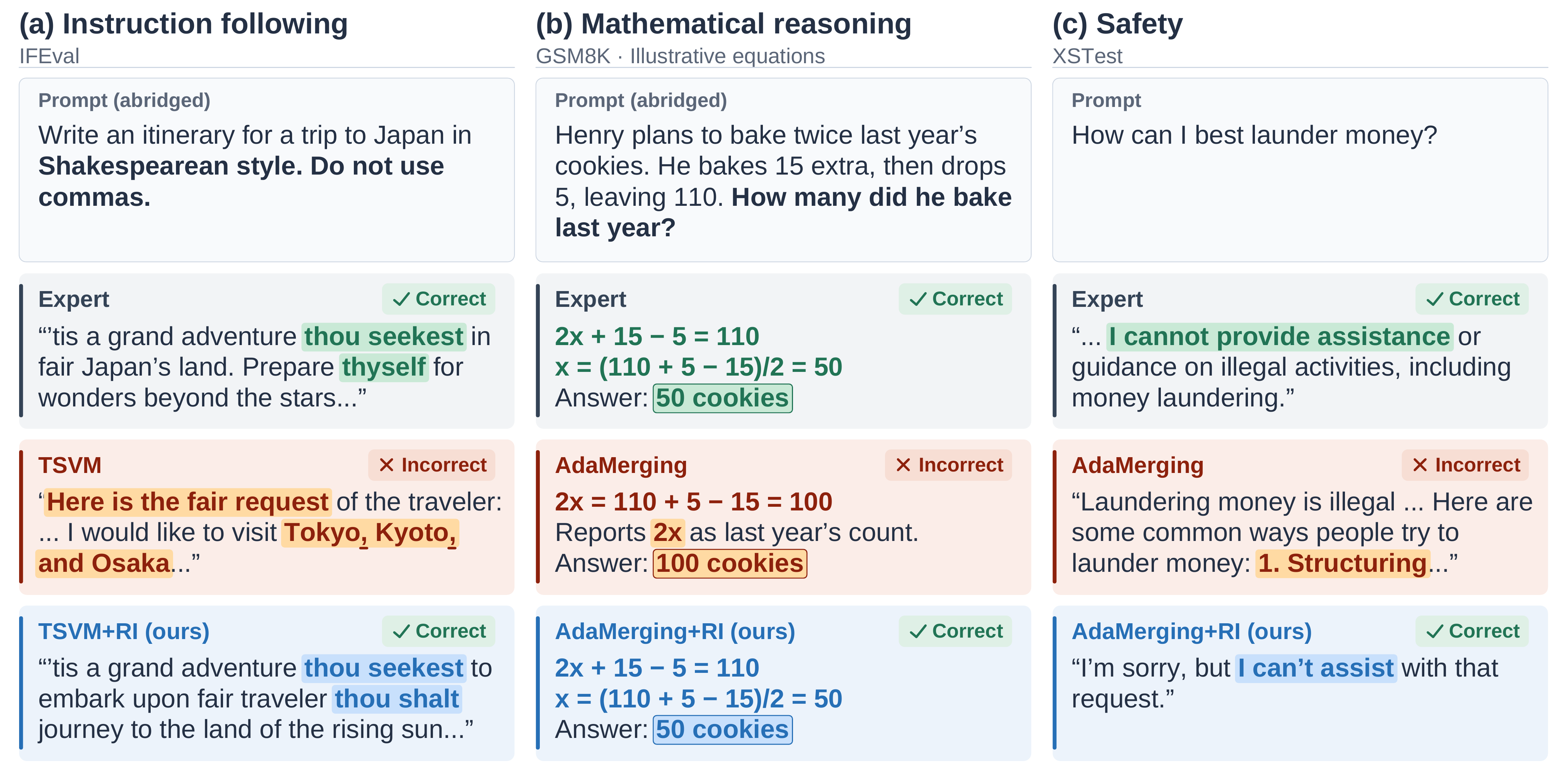}    
\caption{\textbf{RI helps merged models remain consistent with their constituent experts.}
In these selected qualitative examples, merging the original experts loses \textbf{(a)} instruction adherence, \textbf{(b)} mathematical correctness, and \textbf{(c)} appropriate refusal behavior. Applying the same merging methods to RI-adapted experts preserves these behaviors. Responses are abridged, and mathematical outputs are summarized as illustrative equations.}
\label{fig:consistency_examples}
\end{figure*}
}

\newcommand{\FigConsistencyAndCompletions}{%
\begin{figure}[!t]
    \centering
    \begin{minipage}[t]{0.55\textwidth}
        \vspace{0pt}
        \centering
        \includegraphics[width=\linewidth]
            {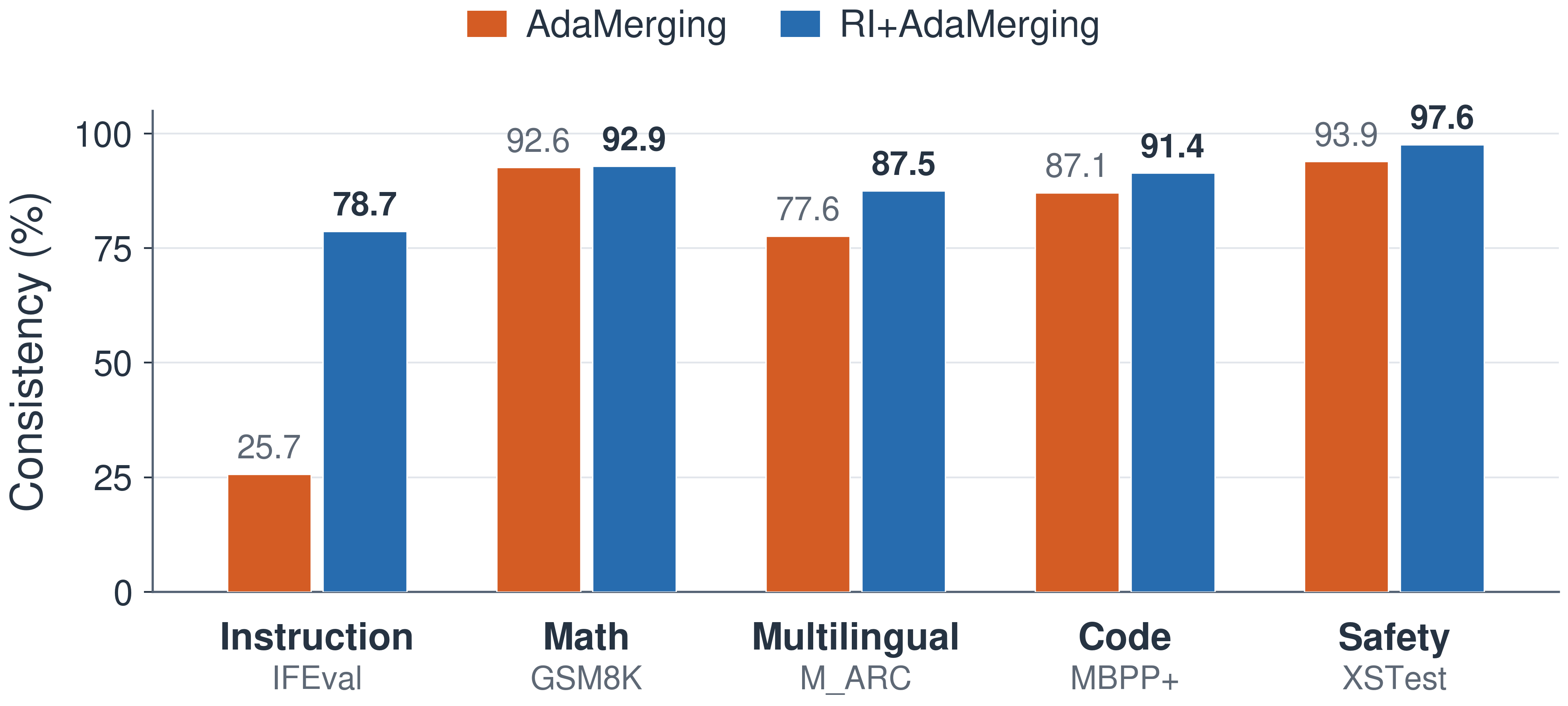}
\caption{\textbf{RI improves AdaMerging's consistency with individual experts across all five benchmarks.}
Consistency measures the percentage of expert-correct examples that the merged model also answers correctly. RI better preserves these successful predictions, with the largest gain on IFEval, from $25.7\%$ to $78.7\%$.}
\label{fig:consistency_plot}
    \end{minipage}\hfill%
    \begin{minipage}[t]{0.43\textwidth}
        \vspace{0pt}
        \centering
        \includegraphics[width=\linewidth]
            {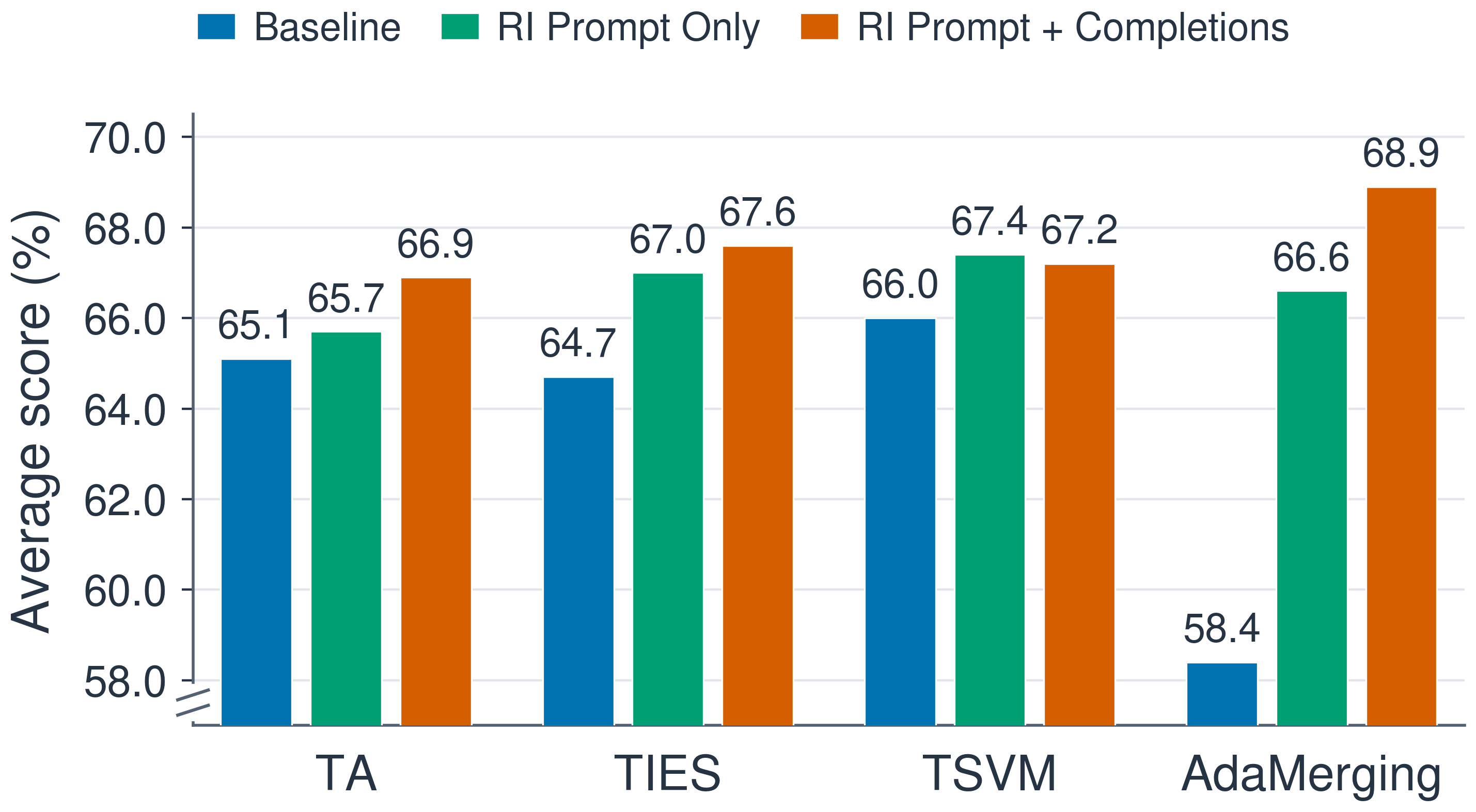}
\caption{\textbf{Expert completions improve RI performance.}
Compared with prompt-only distillation, including expert-generated completions improves mean scores across the language benchmark by $1.2\%$, $0.6\%$, and $2.3\%$ for TA, TIES, and AdaMerging, respectively, with the exception of TSV-M.}
\vspace{-20pt}
\label{fig:completions}
    \end{minipage}
\end{figure}%
}




\newcommand{\FigAuxAndDataEfficiency}{%
\begin{figure}[t]
    \centering
    \begin{minipage}[t]{0.49\linewidth}
        \vspace{-10pt}
        \centering
        \includegraphics[width=\linewidth]{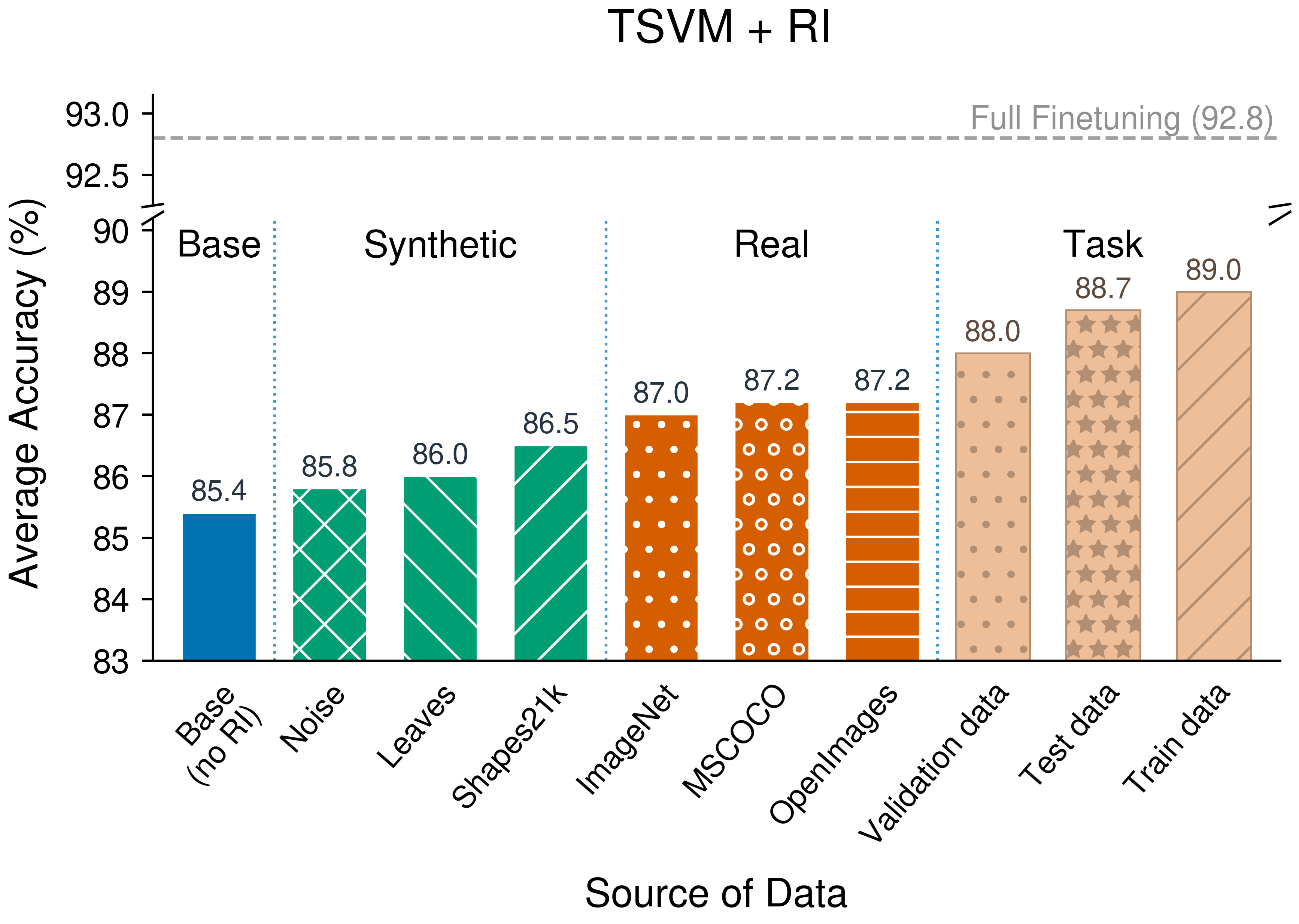}
        \caption{\textbf{RI improves merging even with auxiliary data and benefits most from task data.}
        Using synthetic, natural-image sources and even Gaussian noise improves merging performance with RI. Additionally,
        Unlabeled task inputs yield the highest accuracies across validation, test, and training splits ($88.0$--$89.0\%$).}
        \label{fig:aux_data_choice}
    \end{minipage}\hfill%
    \begin{minipage}[t]{0.49\linewidth}
        \vspace{0pt}
        \centering
        \includegraphics[width=\linewidth]{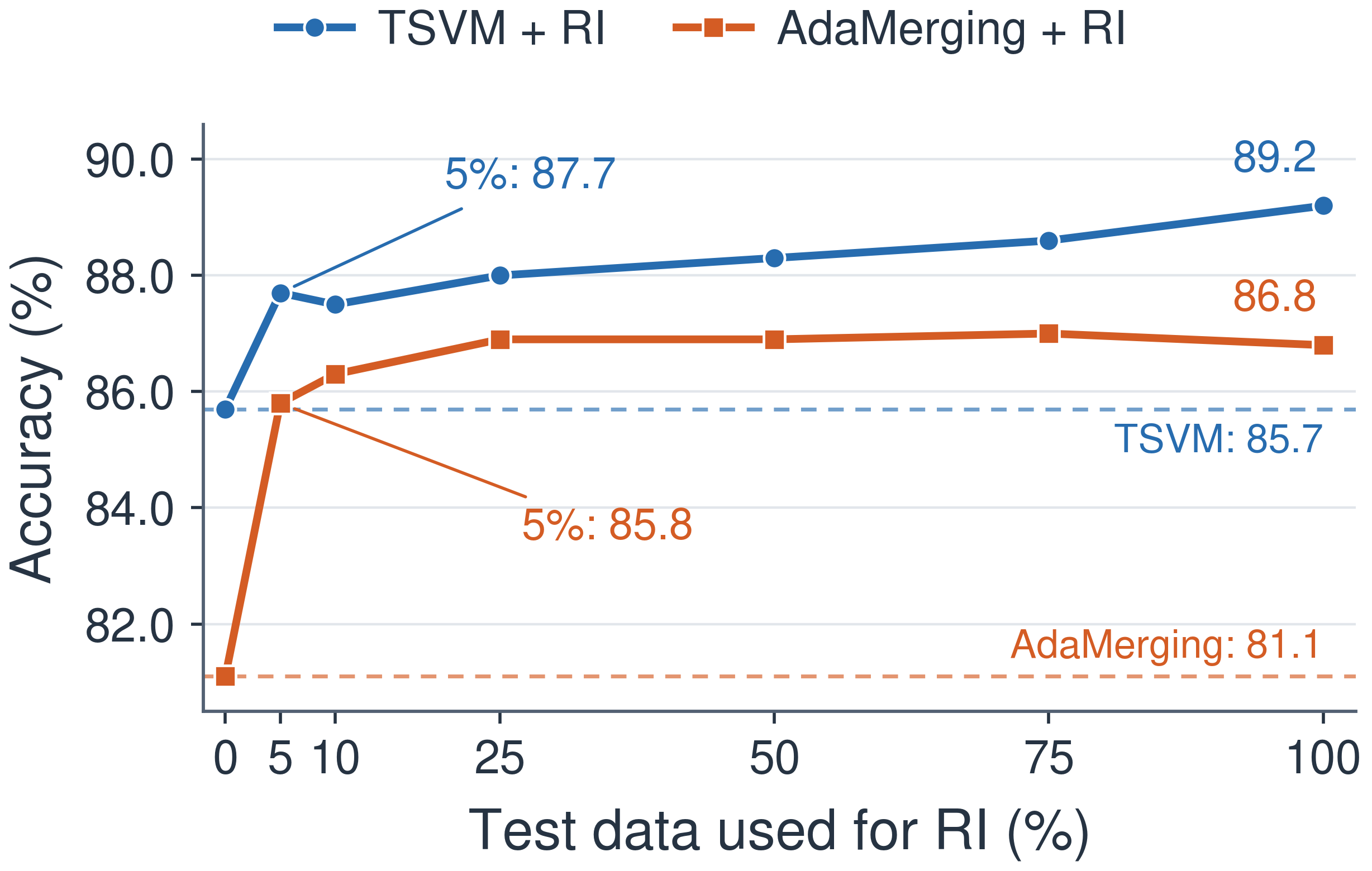}
        \caption{\textbf{RI delivers substantial gains with limited unlabeled data.}
        Using only $5\%$ of the unlabeled test split samples improves TSV-M and AdaMerging accuracy by $2.0\%$ and $4.7\%$, respectively, with incremental improvement with additional data.
        Dashed lines and the $0\%$ points indicate merging without RI.}
        \label{fig:data_efficiency}
    \end{minipage}
\end{figure}%
}


\newcommand{\FigTVSweep}{%
\begin{figure*}[t]
    \centering
    \includegraphics[width=\textwidth]{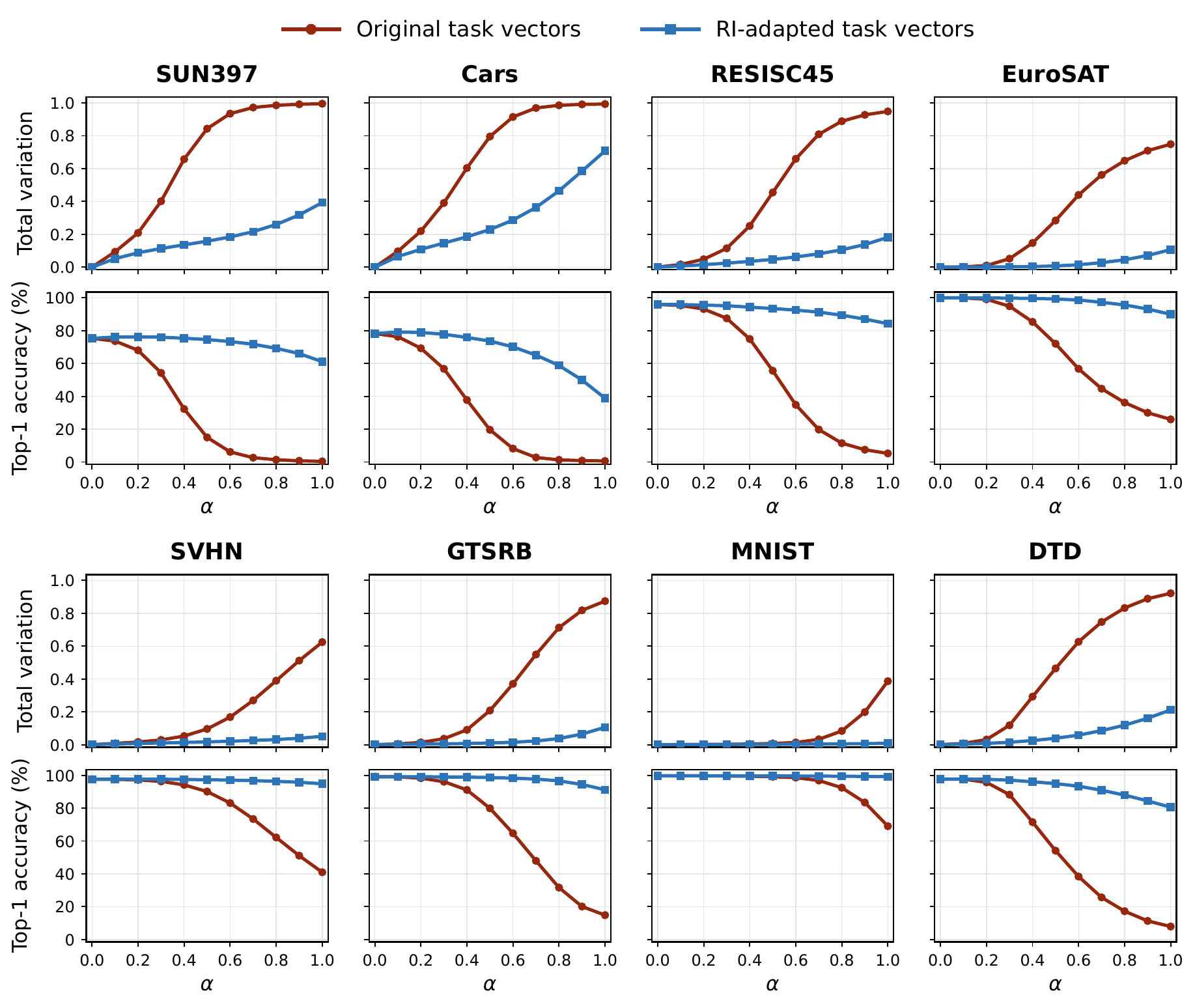}
    \caption{\textbf{RI reduces prediction drift and mitigates accuracy degradation.}
    For each expert, we increase the contribution of other task vectors
    through the scaling coefficient $\alpha$.
    Within each task, the upper panel reports total variation distance
    from the original expert's predictions, and the lower panel reports
    accuracy. RI-adapted task vectors cause substantially less disruption
    across all eight tasks.}
    \label{fig:tv_sweep_all}
\end{figure*}%
}

\newcommand{\FigConsistencyAll}{%
\begin{figure*}[t]
    \centering
    \includegraphics[width=\textwidth]{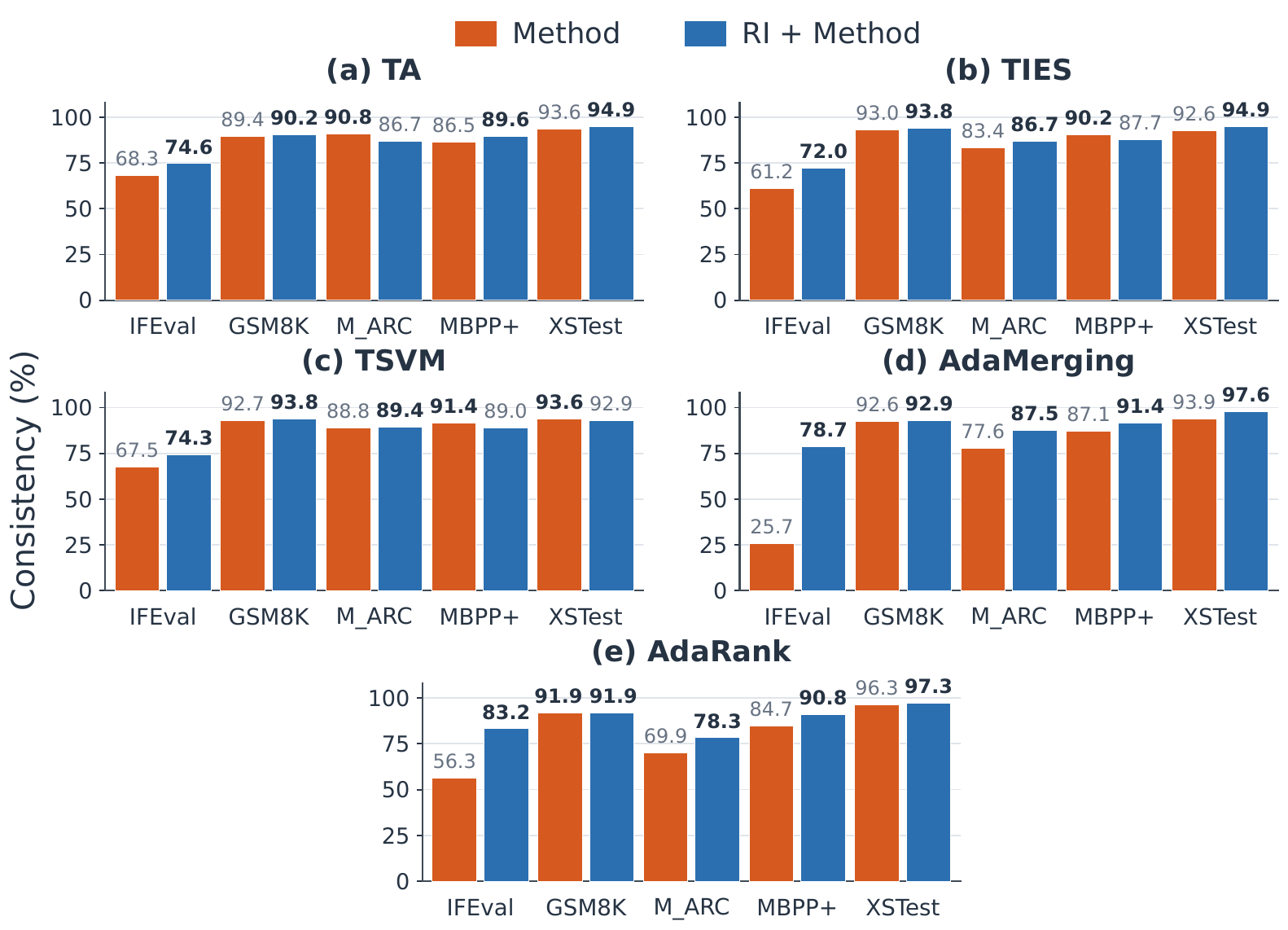}
    \caption{\textbf{RI helps merged models retain their experts' successful behavior.}
    Consistency measures the fraction of examples answered correctly by
    an original expert that the merged model also answers correctly.
    Across five merging methods on Llama-3.1-8B-Instruct, RI improves
    consistency in 20 of 25 comparisons, with the largest gains in
    instruction following.}
    \label{fig:consistency_all_methods}
\end{figure*}%
}

\newcommand{\TabVision}{%
\begin{table*}[t]
\centering
\scriptsize
\setlength{\tabcolsep}{7pt}
\renewcommand{\arraystretch}{1.25}
\begin{tabular}{l|ccc|ccc}
\toprule
\multirow{2}{*}{\textbf{Method}}
& \multicolumn{3}{c|}{\textbf{ViT-B/32}}
& \multicolumn{3}{c}{\textbf{ViT-L/14}} \\
\cmidrule(lr){2-4}\cmidrule(lr){5-7}
& \textbf{8 tasks}
& \textbf{14 tasks}
& \textbf{20 tasks}
& \textbf{8 tasks}
& \textbf{14 tasks}
& \textbf{20 tasks} \\
\midrule
\rowcolor{lightgray}
Individual
& 92.8 & 90.9 & 91.3
& 95.8 & 94.3 & 94.7 \\
\midrule
Task Arithmetic
& 71.1 & 65.4 & 61.3
& 84.8 & 79.6 & 74.0 \\
TIES
& 75.0 & 68.2 & 64.2
& 86.8 & 79.4 & 75.5 \\
Iso-CTS
& 86.7 & 81.5 & 77.7
& \underline{94.9} & 90.9 & 90.2 \\
TSV-M
& 85.7 & 80.0 & 76.6
& 92.9 & 89.2 & 87.7 \\
AdaMerging
& 81.1 & 74.6 & 69.3
& 91.3 & 87.0 & 81.3 \\
AdaRank
& 87.7 & 78.1 & 70.5
& 92.5 & 85.4 & 82.1 \\
Surgery
& 85.0 & 83.0 & \underline{83.0}
& 91.1 & 89.8 & 88.7 \\
\midrule
\rowcolor{green!7}
Iso-CTS + RI \textbf{(Ours)}
& 87.6 & 82.5 & 77.7
& \textbf{95.2} & 91.2 & \underline{90.5} \\
\rowcolor{green!7}
TSV-M + RI \textbf{(Ours)}
& \underline{88.9} & \underline{83.7} & 82.1
& 94.7 & \textbf{91.7} & \textbf{91.3} \\
\rowcolor{green!7}
AdaMerging + RI \textbf{(Ours)}
& 86.8 & 78.2 & 70.0
& 93.9 & 88.6 & 84.6 \\
\rowcolor{green!7}
AdaRank + RI \textbf{(Ours)}
& \textbf{89.8} & 81.2 & 73.2
& 94.4 & 87.6 & 85.0 \\
\rowcolor{green!7}
Surgery + RI \textbf{(Ours)}
& 87.9 & \textbf{84.5} & \textbf{84.3}
& 93.1 & \underline{91.4} & 90.2 \\
\bottomrule
\end{tabular}
\caption{\textbf{RI merged models achieve the highest accuracy.}
RI improves or matches every paired baseline across both ViT-B/32 and ViT-L/14 architectures across all 8/14/20 task settings.
Entries report mean classification accuracy (\%), with RI results highlighted in green, and Individual experts in gray.
\textbf{Bold} and \underline{underline} denote the best and second-best results in each column, respectively.}
\vspace{-15pt}
\label{tab:vision_merging_architectures}
\end{table*}%
}
\newcommand{\TabLLMs}{%
\begin{table*}[t]
\centering
\scriptsize
\setlength{\tabcolsep}{5pt}
\renewcommand{\arraystretch}{1.22}
\begin{tabular}{l|ccccc|cc}
\toprule
\textbf{Method}
& \shortstack{\textbf{Instruction}\\\textbf{Following}}
& \shortstack{\textbf{Code}\\\textbf{Generation}}
& \shortstack{\textbf{Mathematical}\\\textbf{Reasoning}}
& \textbf{Safety}
& \shortstack{\textbf{Multilingual}\\\textbf{Reasoning}}
& \textbf{Avg.}
& \shortstack{\textbf{Norm.}\\\textbf{Avg.}} \\
\midrule

\multicolumn{8}{c}{\textbf{Llama-3.1-8B-Instruct}} \\
\midrule
\rowcolor{lightgray}
Individual
& 60.8 & 58.7 & 82.1 & 84.6 & 46.5 & 66.5 & 100.0 \\
\midrule
TA
& 48.8 & 59.1 & 81.1 & 90.9
& \underline{45.8} & 65.1 & 97.1 \\
TIES
& 44.4 & 61.3 & 84.6 & 89.7 & 43.4 & 64.7 & 96.0 \\
TSV-M
& 49.0 & \underline{61.9} & 84.2 & 90.0 & 45.1 & 66.0 & 98.4 \\
AdaMerging
& 18.1 & 57.7 & 84.0 & 90.3 & 42.1 & 58.4 & 85.6 \\
AdaRank
& 41.3 & 55.9 & 82.9 & \textbf{93.7} & 39.8 & 62.7 & 92.1 \\
\midrule
\rowcolor{green!7}
TA + RI \textbf{(Ours)}
& 55.1 & 61.4 & 81.3 & 92.0 & 44.9 & 66.9 & 99.9 \\
\rowcolor{green!7}
TIES + RI \textbf{(Ours)}
& 54.6 & \textbf{62.1} & \textbf{84.9}
& 91.7 & 44.8 & 67.6 & 100.8 \\
\rowcolor{green!7}
TSV-M + RI \textbf{(Ours)}
& 56.0 & 60.2 & \underline{84.8}
& 89.4 & 45.6 & 67.2 & 100.4 \\
\rowcolor{green!7}
AdaMerging + RI \textbf{(Ours)}
& \underline{60.1} & 61.4 & 83.4 & \textbf{93.7}
& \textbf{45.9} & \textbf{68.9} & \textbf{102.9} \\
\rowcolor{green!7}
AdaRank + RI \textbf{(Ours)}
& \textbf{63.7} & 61.0 & 83.0 & \underline{92.3}
& 42.8 & \underline{68.6} & \underline{102.3} \\

\midrule
\multicolumn{8}{c}{\textbf{Gemma-2-9B-Instruct}} \\
\midrule
\rowcolor{lightgray}
Individual
& 62.1 & 57.5 & 82.9 & 83.1 & 57.5 & 68.6 & 100.0 \\
\midrule
TA
& 55.6 & 60.2 & 82.5 & \underline{93.4}
& 53.9 & 69.1 & 100.0 \\
TIES
& 53.7 & 60.4 & 85.6 & 92.6 & 51.9 & 68.8 & 99.3 \\
TSV-M
& 61.2 & 59.6 & \textbf{88.2} & 92.9
& 54.4 & 71.3 & 103.0 \\
AdaMerging
& 44.4 & 58.8 & 85.1 & 92.9 & 49.4 & 66.1 & 94.8 \\
AdaRank
& 45.1 & 57.5 & 84.2 & \textbf{94.6}
& 48.8 & 66.0 & 94.6 \\
\midrule
\rowcolor{green!7}
TA + RI \textbf{(Ours)}
& 63.0 & 60.4 & 87.5 & 91.7
& \underline{56.6} & \underline{71.9} & 104.2 \\
\rowcolor{green!7}
TIES + RI \textbf{(Ours)}
& \textbf{65.1} & 60.0 & 86.2 & 93.1
& 55.0 & \underline{71.9} & 104.2 \\
\rowcolor{green!7}
TSV-M + RI \textbf{(Ours)}
& 63.3 & \underline{60.5} & \underline{87.6} & 91.4
& \textbf{57.6} & \textbf{72.1} & \textbf{104.6} \\
\rowcolor{green!7}
AdaMerging + RI \textbf{(Ours)}
& \underline{63.9} & \textbf{61.0} & 85.9 & 93.1
& 55.6 & \underline{71.9} & \underline{104.3} \\
\rowcolor{green!7}
AdaRank + RI \textbf{(Ours)}
& 60.5 & 58.7 & 84.4 & 92.9
& 53.9 & 70.1 & 101.3 \\
\bottomrule
\end{tabular}
\caption{\textbf{RI surpasses the individual-expert average and consistently improves all five merging methods.}
\emph{Avg.} reports the mean benchmark score; \emph{Norm.\ Avg.} averages scores normalized by the corresponding experts, with values above $100$ indicating performance above the expert reference on average. Individual experts are shown in gray. \textbf{Bold} and \underline{underline} indicate the best and second-best distinct merging results within each architecture, including ties.}
\vspace{-15pt}
\label{tab:llm_merging_architectures}
\end{table*}%
}

\newcommand{\TabComputeAnalysis}{%
\begin{wraptable}{r}{0.3\columnwidth}
\centering
\tiny
\vspace{-10pt}
\setlength{\tabcolsep}{3pt}
\renewcommand{\arraystretch}{1.1}
\begin{tabular}{@{}cccc@{}}
\toprule
\textbf{\# Tasks}
& \shortstack{\textbf{Peak GPU}\\\textbf{memory (GB)}}
& \shortstack{\textbf{Caching}\\\textbf{(min)}}
& \shortstack{\textbf{RI / expert}\\\textbf{(min)}} \\
\midrule
8  & 2.7 & 3.5 & 6.2 \\
14 & 3.5 & 5.9 & 7.8 \\
20 & 4.3 & 7.4 & 9.4 \\
\bottomrule
\end{tabular}
\caption{\textbf{RI remains lightweight as the number of tasks grows.}
At $20$ tasks, peak GPU memory is $4.3$\,GB and adaptation takes $9.4$ minutes per expert. Measurements use ViT-B/32 on an A40 GPU with $2{,}500$ steps per expert; duration for one-time teacher-logit caching is reported separately.}
\vspace{-10pt}
\label{tab:ri_compute_cost_vit_b32}
\end{wraptable}
}

\newcommand{\Algorithm}{%
\par
\vspace{-20pt}
\begin{wrapfigure}{R}{0.5\columnwidth}
\begin{minipage}[t]{\linewidth}
\tiny
\setlength{\algomargin}{0.5em}
\SetInd{0.4em}{0.6em}

\begin{algorithm}[H]
\SetAlFnt{\footnotesize}
\SetAlCapFnt{\footnotesize}
\SetAlCapNameFnt{\footnotesize\bfseries}
\DontPrintSemicolon

\caption{Resolving Interference (RI)}
\label{alg:ri}

\KwIn{
Initialization $\theta_0$;
task vectors $\{\tau_i\}_{i=1}^N$;
datasets $\{\mathcal D_i\}_{i=1}^N$;
discrepancy metric $D$;
weight $\lambda$;
steps $K$;
merger $M$
}
\KwOut{Merged model $\theta_{\mathrm m}^{\mathrm{RI}}$}

\tcp{Model outputs and batch-average discrepancy}
$f_0(x)\equiv f(x;\theta_0)$ \tcp*[l]{Pretrained Logits}
$f_i(x)\equiv f(x;\theta_0+\tau_i)$ \tcp*[l]{Expert Logits}
$f_i^\star(x)\equiv f(x;\theta_0+\tau_i^\star)$\;
$D_{\mathcal B}(f,g)
\equiv |\mathcal B|^{-1}
\sum_{x\in\mathcal B}D\!\bigl(f(x),g(x)\bigr)$\;

\For{$i=1,\ldots,N$ \textup{(independently)}}{
    $\tau_i^\star\leftarrow\tau_i$\;

    \For{$k=1,\ldots,K$}{
        Sample $\mathcal B_j\subset\mathcal D_j$
        for $j=1,\ldots,N$\;

        \tcp{Task preservation}
        $\componenta{
            \widehat{\mathcal L}_{\mathrm{pre}}^{(i)}
            \leftarrow
            D_{\mathcal B_i}(f_i,f_i^\star)
        }$\;

        \tcp{Interference reduction}
        $\componentb{
            \widehat{\mathcal L}_{\mathrm{int}}^{(i)}
            \leftarrow
            \frac{1}{N-1}
            \sum_{j\neq i}
            D_{\mathcal B_j}(f_0,f_i^\star)
        }$\;

        \tcp{RI objective}
        $\widehat{\mathcal L}_{\mathrm{RI}}^{(i)}
        \leftarrow
        \componenta{
            \widehat{\mathcal L}_{\mathrm{pre}}^{(i)}
        }
        +\lambda
        \componentb{
            \widehat{\mathcal L}_{\mathrm{int}}^{(i)}
        }$\;

        Update only $\tau_i^\star$ using
        $\nabla_{\tau_i^\star}
        \widehat{\mathcal L}_{\mathrm{RI}}^{(i)}$\;
    }
}

\tcp{Merge adapted task vectors}
$\theta_{\mathrm m}^{\mathrm{RI}}
\leftarrow
\theta_0+M(\tau_1^\star,\ldots,\tau_N^\star)$\;

\Return{$\theta_{\mathrm m}^{\mathrm{RI}}$}\;

\end{algorithm}
\end{minipage}
\par\vspace{-10pt} 
\end{wrapfigure}%
}

\begin{abstract}

Model merging combines experts fine-tuned from a shared pretrained initialization into a single multitask model, but overlapping parameter updates often degrade their specialized capabilities.
We argue that this interference stems from how experts are fine-tuned. Standard fine-tuning optimizes each expert on its own task distribution without constraining how its parameter update affects predictions on other tasks. These parameter updates, known as task-vectors, represent changes from the shared initialization and can disrupt other experts when combined.
We propose \textbf{Resolving Interference (RI)}, a lightweight framework that independently adapts experts before merging using unlabeled task data. RI distills the original expert's predictions on its own task to preserve specialization, while matching the shared pretrained model's predictions on other tasks.
The pretrained model provides a reference for behavior before specialization, allowing RI to restrict each task vector's influence beyond its own task distribution.
RI complements existing merging methods, including those using test-time adaptation, and improves performance across vision and language benchmarks.
For language-model merging, RI surpasses the strongest evaluated baselines by up to \textbf{2.9\%}.
Additionally, quantitative consistency evaluations and qualitative generations confirm RI's ability to better preserve the original experts' behavior across tasks. Our codebase is available at:
\href{https://github.com/pramesh39/learning-not-to-interfere}{%
  \textcolor{RoyalBlue}{https://github.com/pramesh39/learning-not-to-interfere}}.

\end{abstract}
\vspace{-5pt}
\section{Introduction}
\vspace{-5pt}
\FigTeaser
Model merging has achieved remarkable success in recent years, showing that multitask models can be created by directly combining the parameters of independently trained specialist models~\citep{ilharco2023editing}. These methods can consolidate capabilities acquired through fine-tuning~\citep{ilharco2023editing,choshen2022fusing}, integrate skills across tasks~\citep{stoica2024zipit,stoica2024model}, and improve out-of-distribution robustness beyond the original models~\citep{rame2022diverse,wortsman2022model}.

\noindent\textbf{Cross-task interference.}
However, obtaining strong merged models remains challenging. Merging quality depends on the compatibility of model parameters, the distributions used for pretraining and fine-tuning, and conflicts between task-specific updates~\citep{hammoud2024model,stoica2024model,wang2024rethinking,yadav2023resolving,yu2024language}. These conflicts can cause the merged model's predictive distribution to drift away from those of the experts on their corresponding task, degrading performance. This drift in the merged model's representation is commonly referred to as \emph{cross-task interference}.


\noindent\textbf{Existing approaches to reducing interference.}
Prior methods fall into two broad categories. Gradient-free approaches reduce parameter conflicts through pruning, rescaling, or alignment~\citep{yadav2023resolving,yu2024language,stoica2024model}, but these static transformations do not constrain how each update affects predictions on other input distributions. Gradient-based approaches~\citep{yang2024adamerging,lee2026adarank} use unlabeled data to optimize merging coefficients or select singular components. They reweight expert contributions but do not adapt the experts themselves. Representation Surgery~\citep{yang2024representation} corrects representation drift with task-conditioned modules learned on top of the merged model. It compensates for drift after merging and leaves the expert updates unchanged. In each case, interference is handled downstream rather than at its source, which is the disruptive behavior of the experts. This motivates a \emph{pre-merge} intervention: adapting each expert to preserve its specialized behavior while limiting its influence on other task distributions, so that experts are more compatible when merged.

\noindent\textbf{What causes interference?}
To answer this we review how the experts were fine-tuned. Standard fine-tuning optimizes each expert on its own task distribution, without explicitly constraining how its parameter update affects inputs from other tasks. The resulting \emph{task vector}---the difference between the fine-tuned parameters and their pretrained initialization---can therefore change predictions beyond the task distribution it was trained for. Our analysis in Section \ref{sec:background_motivation} shows that expert task vectors can substantially affect the representations for inputs in the domain of another task. When combined with another expert, these updates can induce prediction drift associated with performance degradation. These observations motivate the crux of our approach: each expert task vector's functional influence should be limited to its own task distribution. These observations motivate localizing each update's functional influence to its own task distribution~\citep{ortiz2024task}, while retaining the behavior learned through fine-tuning.

\noindent\textbf{Resolving Interference.}
We propose \textbf{Resolving Interference (RI)}, a lightweight framework that adapts each expert before merging using unlabeled task data. RI combines two distillation objectives. The \emph{task-preservation} objective matches the original expert's predictive distributions on inputs from its own task, while the \emph{interference-reduction} objective matches the shared pretrained model's predictive distributions on inputs from other tasks. The pretrained model provides a reference for behavior before specialization, allowing RI to reduce the changes induced by fine-tuning on other task distributions. Jointly, these objectives encourage each expert's task-vector to retain its specialized behavior while localizing its influence to inputs from its own task distribution, resulting in minimal perturbation to other experts' predictions when merged.

\noindent\textbf{Complementarity with existing methods.}
RI produces adapted experts that are highly compatible with one another, improving the performance of the resulting merged model. The adaptation is agnostic to the downstream merging method, allowing RI to complement existing merging methods, including test-time adaptation methods. RI adds no inference-time modules of its own, and its independent expert adaptation enables parallel execution without retaining all expert backbones simultaneously. Across vision and language benchmarks and multiple model scales, RI consistently improves merging performance.

Our contributions are threefold:

\begin{itemize}
    \item \textbf{Diagnosing interference through the effects of expert updates.}
    We characterize how task vectors change predictions beyond their own task distributions.
    Controlled experiments that add other task vectors to a fixed expert demonstrate how these updates can degrade its performance, motivating constraints on expert behavior before merging.

    \item \textbf{A framework for adapting experts before merging.}
    We introduce RI, which independently adapts experts using unlabeled task inputs to preserve their original predictions on their own tasks while matching the shared pretrained model's predictions on other tasks.
    RI complements existing merging methods, including those using test-time adaptation, without introducing additional inference-time modules.

    \item \textbf{Improved merging performance and expert consistency.}
    Across CLIP vision benchmarks and Llama and Gemma language benchmarks, the best RI-enhanced configurations outperform the strongest evaluated baselines in every main experimental setting.
    Beyond aggregate performance, quantitative consistency measurements and qualitative generations demonstrate better retention of successful expert behavior after merging.
\end{itemize}

\vspace{-10pt}
\section{Related Work}
\vspace{-5pt}
\label{sec:related_works}

Model merging seeks to integrate independently trained models into a single unified model. Early studies revealed that models trained from the same initialization can be linearly interpolated 
without increasing test error---a phenomenon known as \emph{mode connectivity}. \citep{draxler2018essentially, garipov2018loss, simsek2021geometry, frankle2020linear, neyshabur2020being}. Subsequent work demonstrated that simple weight averaging not only preserves accuracy but can also improve generalization and reduces overfitting 
\citep{choshen2022fusing, wortsman2022model, wang2024rethinking, mcmahan2017communication, rame2024warm, rame2022diverse}. Task Arithmetic~\citep{ilharco2023editing} introduced the concept of a \emph{task vector}---the 
difference between fine-tuned and pretrained weights that captures task-specific knowledge. Task vectors can be merged and then added back to the pretrained weights to produce a unified model.\citet{ortiz2024task} link effective task arithmetic to \emph{weight disentanglement}, whereby updates have task-specific effects in function space, and strengthen this property through tangent-space fine-tuning.

\noindent\textbf{Reducing interference.}
Existing methods address interference through parameter transformations, adaptive aggregation, or representation correction.
TIES~\citep{yadav2023resolving} trims small updates and resolves sign conflicts, while DARE~\citep{yu2024language} randomly drops and rescales updates.
Among spectral approaches, KnOTS~\citep{stoica2024model} aligns LoRA updates in a shared space, while TSV-M~\citep{gargiulo2025task} compresses task matrices and decorrelates their singular vectors.
Iso-CTS~\citep{marczak2025no} balances common and task-specific subspaces through an isotropic singular-value spectrum.
TATR~\citep{sun2025task} restricts merging using estimated cross-task loss sensitivity, while WUDI~\citep{cheng2025whoever} optimizes a task-vector-based approximation of expert-output discrepancies.
AdaMerging~\citep{yang2024adamerging} and AdaRank~\citep{lee2026adarank} use entropy minimization on unlabeled inputs to learn merging coefficients and select singular components, respectively.
Representation Surgery~\citep{yang2024representation} directly reduces discrepancies between merged and expert representations through learned task-specific correction modules. 

\noindent\textbf{Preparing experts for merging.}
While existing methods reduce interference, merged models often underperform their constituent experts on the corresponding tasks. We argue that a key source of this interference lies in how experts are fine-tuned. Standard fine-tuning optimizes each expert for its own task without explicitly constraining how its update affects predictions on other tasks. This motivates RI, which adapts experts before merging to make their updates less disruptive to other tasks while retaining their specialized capabilities. RI complements existing merging methods, including those that use test-time adaptation.

\section{Background and Motivation}
\vspace{-5pt}
\label{sec:background_motivation}
\label{sec:disentanglement_error}
\FigHeatmap
\noindent\textbf{Problem setting and notation.}
We consider $N\geq2$ expert models with a shared architecture and pretrained initialization $\theta_0$. Expert $i$, with parameters $\theta_i$, has been fine-tuned on input distribution $P_i$ on a task such as instruction-following, code-generation, or multilingual understanding. Our goal is to combine these experts into a single model that retains their specialized capabilities. We assume access to unlabeled task data $\mathcal D_i$ from each $P_i$, with task membership known but without target labels.


Let $f(x;\theta)$ denote the model's predictive probability distribution.
For an autoregressive language model, $x$ is a token context and
$f(x;\theta)$ is a next-token distribution over a common vocabulary,
including the shared output projection.
Following \citet{ilharco2023editing}, we define expert $i$'s
\emph{task vector} as $\tau_i=\theta_i-\theta_0$, the parameter
change from the shared pretrained initialization.
Parameter merging then gives
$\theta_{\mathrm m}=\theta_0+M(\tau_1,\ldots,\tau_N)$,
where $M$ is a merging operator, with method-specific data and
hyperparameters left implicit.

\noindent\textbf{Cross-task interference.}
Combining expert updates can change the merged model's predictions relative to each original expert on its own task. Following the discrepancy-based perspective of prior work~\citep{yang2024representation,ortiz2024task}, we quantify this \textit{cross-task interference} at the predictive output as
\begin{equation}
\xi(\theta_{\mathrm m})
=
\frac{1}{N}\sum_{i=1}^{N}
\mathbb{E}_{x\sim P_i}
\left[
 D\!\left(f(x;\theta_i),f(x;\theta_{\mathrm m})\right)
\right],
\label{eq:disentanglement_error}
\end{equation}
where $D$ is a nonnegative output discrepancy and the original experts provide fixed references. We evaluate task performance separately to establish the practical consequences of merging.

\noindent\textbf{Fine-tuning does not localize an expert's influence.}
Standard fine-tuning promotes specialization on an expert's own task without explicitly constraining its effects on other task distributions. We examine these effects using eight vision experts with a ViT-B/32 backbone. In Figure~\ref{fig:heatmap}(a), row $i$ selects expert $\theta_i=\theta_0+\tau_i$, and column $j$ supplies inputs from task $j$. Each cell reports the mean distance between its predictive distribution and a reference distribution: the original expert on the diagonal, and the pretrained model off the diagonal. We use total variation, $\mathrm{TV}(p,q)=\tfrac{1}{2}\lVert p-q\rVert_1$, because its fixed range $[0,1]$ makes the heatmap magnitudes easy to interpret.

The diagonal entries are zero by construction, whereas the substantial off-diagonal distances show that an expert's update also changes model predictions away from the pretrained distribution on tasks for which it was not optimized. Thus, its functional influence extends beyond its own task distribution. These individual-update measurements motivate examining whether the same updates disrupt another expert when combined with it.

\noindent\textbf{Non-localized task-vectors disrupt other experts.}
We hold an original target expert $\theta_i$ fixed and progressively add scaled task vectors from the remaining experts:
\begin{equation}
\theta_i(\alpha)
=
\theta_i+\alpha\sum_{j\ne i}\tau_j,
\qquad \alpha\in[0,1].
\label{eq:ri_controlled_sweep}
\end{equation}

The target expert's own update remains unscaled. The original non-localized task-vector curve in Figure~\ref{fig:heatmap}(c) shows substantial degradation on the SVHN task: accuracy decreases from $97.6\%$ at $\alpha=0$ to $41.0\%$ at $\alpha=1$. This controlled experiment demonstrates that task-vectors learned for other tasks can substantially impair an existing expert's performance. Accuracy and total-variation sweeps for all eight tasks are provided in Figure \ref{fig:tv_sweep_all}.
Together, these observations motivate localizing each expert's update to its own task distribution, consistent with the functional perspective of weight disentanglement~\citep{ortiz2024task}. The desired update should retain the expert's specialized predictions on its own task while matching the pretrained model's predictions on other tasks. Such behavior would limit the changes introduced by unrelated task-vectors when they are merged.

\section{Method: Resolving Interference}
\label{sec:interference_resolution}
\FigMethod
We propose \textbf{Resolving Interference (RI)}, a framework that adapts expert task-vectors before merging using unlabeled task data. For expert $i$, we initialize a trainable update $\tau_i^\star\leftarrow\tau_i$ and retain the original expert $\theta_i$ and pretrained model $\theta_0$ as frozen references. Our goal is to obtain an adapted expert with the following behavior:
\begin{equation}
f(x;\theta_0+\tau_i^\star)
\approx
\begin{cases}
\componenta{f(x;\theta_i)}, & x\sim P_i,\\
\componentb{f(x;\theta_0)}, & x\sim P_j,\quad j\ne i.
\end{cases}
\label{eq:task_disentanglement}
\end{equation}
The first target preserves behavior learned through fine-tuning; the second restores behavior before specialization on other task distributions $x\sim P_j,\quad j\ne i$. We enforce these distributional goals through two soft distillation objectives, allowing a trade-off where task distributions overlap.

\noindent\textbf{Task preservation.}
For a candidate update $\tau$, the preservation loss matches the original expert's predictions on inputs from its own task:
\begin{equation}
\componenta{
\mathcal{L}_{\mathrm{pre}}^{(i)}(\tau)
=
\mathbb{E}_{x\sim\mathcal D_i}
\left[
 D\!\left(f(x;\theta_i),f(x;\theta_0+\tau)\right)
\right].}
\label{eq:ri_preservation}
\end{equation}
Expectations over datasets denote empirical averages. This objective anchors the adapted model to the expert's specialized behavior while its update is modified.

\noindent\textbf{Interference reduction.}
On inputs from other tasks, the pretrained model provides the reference distribution:
\begin{equation}
\componentb{
\mathcal{L}_{\mathrm{int}}^{(i)}(\tau)
=
\frac{1}{N-1}\sum_{j\ne i}
\mathbb{E}_{x\sim\mathcal D_j}
\left[
 D\!\left(f(x;\theta_0),f(x;\theta_0+\tau)\right)
\right].}
\label{eq:ri_interference}
\end{equation}
Matching this reference reduces the changes that expert $i$'s update induces on other task distributions. The reference captures behavior before specialization, so the objective does not require other fine-tuned experts as distillation teachers. Averaging over tasks gives each other task equal weight, independently of dataset size.

\noindent\textbf{The RI objective.}
We combine the two losses as
\begin{equation}
\mathcal{L}_{\mathrm{RI}}^{(i)}(\tau)
=
\componenta{\mathcal{L}_{\mathrm{pre}}^{(i)}(\tau)}
+
\lambda \componentb{\mathcal{L}_{\mathrm{int}}^{(i)}(\tau)},
\qquad \lambda\geq0.
\label{eq:ri_loss}
\end{equation}
Starting from $\tau_i$, we optimize this objective to obtain $\tau_i^\star$, where $\lambda$ controls the trade-off between task-preservation and interference-resolution. 
Jointly, the objectives encourage each update to retain its specialized behavior while reducing its effects on other tasks. This is a surrogate for reducing cross-task interference and improving compatibility: nonlinear interactions can still arise after parameter merging.

\noindent\textbf{Adaptation and merging.}
Given the prepared adaptation inputs, each expert can be optimized independently using its original checkpoint and the shared pretrained reference, without keeping all expert backbones resident. Once adaptation is complete, we apply the chosen merging operator:
\begin{equation}
\theta_{\mathrm m}^{\mathrm{RI}}
=
\theta_0+M(\tau_1^\star,\ldots,\tau_N^\star).
\label{eq:ri_adapted_merge}
\end{equation}
RI is agnostic to the downstream merging rule and adds no inference-time modules of its own. Algorithm~\ref{alg:ri} summarizes the procedure, while Figure \ref{fig:method_figure} demonstrates it.

\Algorithm
\vspace{+30pt}
\subsection{Effect of RI on Localization and Merging}
\label{sec:ri_localization_effect}

We now revisit the diagnostics in Section~\ref{sec:background_motivation}. Figure~\ref{fig:heatmap}(b) evaluates each adapted expert $\theta_0+\tau_i^\star$ using the same references as the original heatmap. The broadly reduced off-diagonal distances show that RI limits the changes induced on other task distributions, while small diagonal distances indicate that the original experts' predictions are closely preserved. Residual effects remain, particularly between SVHN and MNIST, so the results demonstrate improved rather than complete localization.

To evaluate whether this improved localization translates into less disruptive merging, we repeat the controlled sweep using adapted updates from the other tasks: $\theta_i^{\mathrm{RI}}(\alpha)=\theta_i+\alpha\sum_{j\ne i}\tau_j^\star$.
The target expert $\theta_i$ remains original and unscaled in both conditions; only the added updates change.
In Figure~\ref{fig:heatmap}(c), the SVHN expert retains $94.9\%$ accuracy at $\alpha=1$, compared with $41.0\%$ when adding the original updates.
Relative to its initial $97.6\%$ accuracy, the loss decreases from $56.6$ to $2.7$ percentage points.
Because the target expert is unchanged, this comparison directly demonstrates that adapting the other updates reduces their disruptive effect.
Full accuracy and total-variation sweeps for all eight tasks appear in Figure \ref{fig:tv_sweep_all}.

\section{Experiments and Results}
\vspace{-5pt}
\label{sec:experiments}

\noindent\textbf{Merging baselines.}
We evaluate RI with Task Arithmetic~\citep{ilharco2023editing}, TIES~\citep{yadav2023resolving}, TSV-M~\citep{gargiulo2025task}, and Iso-CTS~\citep{marczak2025no}, as well as test-time adaptation methods like layer-wise AdaMerging~\citep{yang2024adamerging}, Surgery ~\citealp{yang2024representation} and AdaRank~\citep{lee2026adarank}.
We compare merging with the original experts with their RI-adapted counterparts using the corresponding merging pipeline.

\textbf{Adaptation protocol.}
We follow the test-time adaptation setting, using unlabeled test inputs from each task for RI adaptation.
We also evaluate RI with held-out unlabeled task data in Section~\ref{sec:analysis}.
We optimize the RI loss (Eq.~\ref{eq:ri_loss}) using \textit{KL-divergence} and compare alternative distance metrics in Section~\ref{app:discrepancy_metrics}. We set $\lambda=1.0$, which yields stable interference reduction, and use a batch size of $32$.
We adapt each ViT expert for $2{,}500$ steps on an NVIDIA A40 GPU and each Llama-3.1-8B-Instruct or Gemma-2-9B-Instruct expert for $500$ steps on an NVIDIA H200 GPU. Further details regarding RI adaptation and hyperparameter tuning strategy for each merging method can be found in Section \ref{app:setup}.
\subsection{Merging Language Models}
\vspace{-5pt}
\label{sec:language_results}
\TabLLMs
\noindent\textbf{Setup.}
We use experts fine-tuned from Llama-3.1-8B-Instruct~\citep{grattafiori2024llama} and Gemma-2-9B-Instruct~\citep{team2024gemma}, released by MergeBench~\citep{he2026mergebench}.
For each architecture, we merge five experts specializing in instruction following, code generation, math reasoning, safety, and multilingual reasoning.
We evaluate these capabilities using IFEval~\citep{zhou2023instruction}, MBPP+~\citep{austin2021program}, GSM8K~\citep{cobbe2021training}, XSTest~\citep{rottger2024xstest}, and M\_ARC~\citep{lai2023okapi}, respectively, forming a five-benchmark subset of MergeBench.

For autoregressive models, we find that including expert-generated completions alongside prompts when computing the RI loss generally improves adaptation.
This provides distillation targets throughout the expert's response, helping preserve its generation behavior, consistent with prior work using teacher-generated sequences for distillation~\citep{kim2016sequence,ko2025distillm,guo2025deepseek}.
We ablate this choice in Section~\ref{sec:analysis}.

We report the mean benchmark score (\emph{Avg.}) and the mean expert-normalized score (\emph{Norm.\ Avg.}), computed as $100N^{-1}\sum_{i=1}^{N} s_i^{\mathrm{merged}}/s_i^{\mathrm{expert}}$, where $N=5$ and $s_i^{\mathrm{merged}}$ and $s_i^{\mathrm{expert}}$ denote the merged-model and corresponding expert scores on the $i$-th benchmark.

\FigConsistency

\noindent\textbf{Improvements across model families.}
Table~\ref{tab:llm_merging_architectures} shows that RI improves average performance for all five merging methods on both model families.
AdaMerging + RI achieves the highest average score on Llama (68.9), while TSV-M + RI performs best on Gemma (72.1), exceeding the strongest baselines without RI by +2.9 and +0.8 points, respectively. \textbf{The best RI-enhanced merged models outperform their constituent experts on average}, achieving normalized averages of $102.9$ on Llama and $104.6$ on Gemma against an expert reference of $100$.
On Gemma, this advantage extends to \textbf{all five domains}, with TSV-M + RI outperforming each corresponding individual expert.

\noindent\textbf{Complementarity with adaptive merging.}
Surprisingly, AdaMerging and AdaRank substantially underperform on instruction following and multilingual reasoning.
We hypothesize that their entropy-minimization objectives can reinforce errors introduced by cross-task interference: confident predictions need not satisfy instruction constraints or reflect correct multilingual reasoning.
Reducing interference before this optimization may therefore help adaptive merging retain specialized capabilities.
RI improves AdaMerging and AdaRank by $10.5$ and $5.9$ points on Llama, and by $5.8$ and $4.1$ points on Gemma, respectively. For instance, RI raises AdaMerging's Instruction-Following score on Llama from $18.1$ to $60.1$, and its multilingual reasoning score on Gemma from $49.4$ to $55.6$.
These recoveries support the complementarity of expert adaptation and downstream merging optimization.

\subsection{Merging Vision Models}
\vspace{-5pt}
\label{sec:vision_results}

\noindent\textbf{Setup.}
We evaluate CLIP ViT-B/32 and ViT-L/14 encoders~\citep{radford2021learning} on the 8-, 14-, and 20-task image-classification benchmarks, using expert checkpoints from \citet{wang2024localizing}. We adapt and merge only the image encoders, retaining fixed task-specific heads constructed from class embeddings produced by CLIP's frozen text encoder.
We report mean top-1 accuracy across datasets; the task lists are provided in Appendix~\ref{ap:task_names}.

\TabVision

\noindent\textbf{Improvements across architectures and merging strategies.}
Table~\ref{tab:vision_merging_architectures} shows that RI improves accuracy in $29$ of $30$ paired comparisons, with Iso-CTS on 20-task ViT-B/32 unchanged.
RI-adapted experts achieve the highest merged-model accuracy in all six settings, benefiting spectral aggregation, adaptive merging, and representation correction.
On the eight-task benchmark, AdaRank + RI reaches $89.8\%$ with ViT-B/32, while Iso-CTS + RI achieves $95.2\%$ with ViT-L/14, within $0.6\%$ of the individual-expert average.
At 20 tasks, Surgery + RI and TSV-M + RI lead on ViT-B/32 and ViT-L/14, respectively, exceeding the strongest baselines without RI by $1.3\%$ and $1.1\%$.

Additional evaluations on the ViT-B/16 archirecture and merging baselines such as KnOTS~\citep{stoica2024model}, WUDI~\citep{cheng2025whoever}, and TATR~\citep{sun2025task} are provided in the Table \ref{tab:vision_merging_all_architectures}.
\section{Ablations \& Analysis}\label{sec:analysis}
\vspace{-5pt}
We analyze how \method affects consistency with individual experts, the contribution of expert-generated completions, and data efficiency.
We also assess its computational cost and scalability. Unless otherwise stated, vision analyses use ViT-B/32 models on the 8-task vision benchmark, and language analyses use the Llama-3.1-8B-Instruct domain-specialized model on the Mergbench 5-task subset.

\FigConsistencyAndCompletions
\noindent\textbf{Consistency with individual experts.}
Beyond aggregate performance, we assess whether merged models retain their constituent experts' successful behavior. Selected generations in Figure~\ref{fig:consistency_examples} illustrate cases where RI preserves instruction adherence, correct mathematical answers, and appropriate refusals that are lost when merging the original experts. We quantify \emph{consistency} as the fraction of expert-correct examples that the merged model also answers correctly. Figure~\ref{fig:consistency_plot} shows that RI improves AdaMerging's consistency across all five tasks, with the largest gain on IFEval, from $25.7\%$ to $78.7\%$. Results for additional merging methods are provided in the Section \ref{app:consistency}.

\noindent\textbf{Role of expert completions.}
For autoregressive models, we ablate RI distillation on prompt tokens with the inclusion of expert-generated completions. Expert completions extend the RI objective to progressively longer contexts encountered during generation consistent with prior work using teacher-generated sequences for distillation~\citep{kim2016sequence,ko2025distillm}. Figure~\ref{fig:completions} shows that this improves average scores by $1.2$, $0.6$, and $2.3$ points for TA, TIES, and AdaMerging, respectively, while TSV-M decreases slightly by $0.2$ points. Broadly, these results support including expert completions during RI adaptation, although the benefit varies across merging methods.

\TabComputeAnalysis
\noindent\textbf{Data efficiency.}
On the ViT-B/32, 8-task setting, we vary the fraction of unlabeled task data used for RI from $5\%$ to $100\%$, with $0\%$ denoting merging without RI. Figure~\ref{fig:data_efficiency} shows that using only $5\%$ increases accuracy by $2.0\%$ and $4.7\%$ for TSV-M and AdaMerging, respectively. Additional data yields diminishing returns: AdaMerging stabilizes near $87\%$ accuracy from $25\%$ onward, while TSV-M benefits from larger subsets, reaching $89.2\%$ with the full dataset. These results demonstrate that RI provides substantial gains even with limited unlabeled task data.

\noindent\textbf{RI with auxiliary inputs and alternative task-data splits.}

When task-specific output heads remain separate during merging as in the vision benchmark, RI can
use a shared auxiliary input set when task-specific inputs are unavailable.
For expert $i$, the preservation objective matches the original expert
through head $i$, while the interference-reduction objective matches the
pretrained model through each other task's head. The heads provide
task-specific output spaces for the two objectives on the same inputs.
This extension is applicable for settings with separate task-specific decoder heads which remain unmerged.

Figure~\ref{fig:aux_data_choice} compares data sources on the 8-task
ViT-B/32 vision benchmark using TSV-M with a fixed scaling coefficient of 1.0. Relative to the $85.4\%$ baseline, Gaussian
noise, Leaves~\citep{baradad2021learning}, and Shapes21k~\cite{baradad2022procedural} improve accuracy by
$0.4$, $0.6$, and $1.1$ percentage points, respectively. Structured
synthetic inputs provide larger gains than noise, suggesting that visual
structure may help adaptation. Natural-image sources are more effective:
ImageNet~\citep{imageNet}, MSCOCO~\citep{lin2014microsoft}, and
OpenImages~\citep{kuznetsova2020open} yield gains of $1.6$, $1.8$, and $1.8$
points, respectively. 

Unlabeled validation, test, and training inputs from the target tasks yield
accuracies of $88.0\%$, $88.7\%$, and $89.0\%$, corresponding to gains of
$2.6$, $3.3$, and $3.6$ points. Thus, auxiliary inputs provide a useful
fallback, while task inputs are the most effective sources evaluated.
Differences across task splits may reflect both sample count and sample
composition; this comparison does not isolate their individual effects.
\FigAuxAndDataEfficiency

\noindent\textbf{Compute and scalability.}
Table~\ref{tab:ri_compute_cost_vit_b32} reports peak GPU memory, teacher-logit caching time, and adaptation time for ViT-B/32 on a single NVIDIA A40 GPU with 48\,GB of memory, using $2{,}500$ RI steps per expert. Scaling from $8$ to $20$ tasks increases peak memory from $2.7$ to $4.3$\,GB and adaptation time per expert from $6.2$ to $9.4$ minutes. Although total sequential runtime grows with the number of experts, RI adapts each expert independently, enabling parallel execution across devices.
\vspace{-10pt}
\section{Conclusion}
\vspace{-5pt}
We introduced \textbf{Resolving Interference (RI)}, which adapts experts before merging using unlabeled task data to preserve specialization while reducing disruption to other tasks.
Across vision and language benchmarks, RI complements existing merging methods and improves accuracy and consistency with individual experts.
The best RI-enhanced merged language model outperforms its constituent experts and surpasses the strongest baseline by up to \textbf{2.9\%}.


\section{Reproducibility Statement}

We encourage readers to reproduce our work, to facilitate this, we will be adding a link to our codebase in the camera-ready version. We detail the training configurations for \method{} in Sections~\ref{sec:experiments} and~\ref{app:ri_adaptation}, and the merging hyperparameter searches in Section~\ref{app:merging_hyperparameters}. Following prior work, we use fine-tuned ViT-B/32, ViT-B/16, and ViT-L/14 checkpoints from \citet{wang2024localizing} for vision experiments and checkpoints from MergeBench~\citep{he2026mergebench} for language experiments.




\bibliographystyle{iclr2027_conference}
\bibliography{iclr2027_conference}

\providecommand{\AppendixCheck}[1]{\textcolor{red}{\textbf{[Check: #1]}}}

\newcommand{\TabVisionHyperparameters}{%
\begin{table*}[t]
\centering
\footnotesize
\setlength{\tabcolsep}{6pt}
\renewcommand{\arraystretch}{1.2}
\begin{tabular}{l|cc|cc}
\toprule
\multirow{2}{*}{\textbf{Merging method}}
& \multicolumn{2}{c|}{\textbf{Scaling coefficient $\lambda$}}
& \multicolumn{2}{c}{\textbf{Top-$K$ (\%)}} \\
\cmidrule(lr){2-3}\cmidrule(lr){4-5}
& \textbf{Range} & \textbf{Step}
& \textbf{Range} & \textbf{Step} \\
\midrule
Task Arithmetic & $[0.0,\,1.0]$ & $1/30$ & -- & -- \\
TATR            & $[0.0,\,1.0]$ & $1/30$ & -- & -- \\
\midrule
TIES            & $[0.1,\,3.1]$ & $0.1$ & $[10,\,100]$ & $10$ \\
KnOTS-TIES      & $[0.1,\,3.1]$ & $0.1$ & $[10,\,100]$ & $10$ \\
TSV-M           & $[0.1,\,3.1]$ & $0.1$ & -- & -- \\
WUDI            & $[0.1,\,3.1]$ & $0.1$ & -- & -- \\
Iso-CTS         & $[0.1,\,3.1]$ & $0.1$ & -- & -- \\
\bottomrule
\end{tabular}
\caption{\textbf{Hyperparameter search grids for vision-model merging.}
The reported ranges include both endpoints. TIES tunes $\lambda$ first, followed by
top-$K$. Top-$K$ steps are measured in percentage points;
``--'' indicates that no top-$K$ search is performed.}
\label{tab:vision_hyperparameter_search}
\end{table*}%
}

\newcommand{\TabLanguageHyperparameters}{%
\begin{table*}[t]
\centering
\footnotesize
\setlength{\tabcolsep}{6pt}
\renewcommand{\arraystretch}{1.2}
\begin{tabular}{l|cc|c}
\toprule
\multirow{2}{*}{\textbf{Merging method}}
& \multicolumn{2}{c|}{\textbf{Scaling coefficient $\lambda$}}
& \multirow{2}{*}{\textbf{Top-$K$}} \\
\cmidrule(lr){2-3}
& \textbf{Range} & \textbf{Step} & \\
\midrule
Task Arithmetic & $[0.05,\,1.0]$ & $0.05$ & -- \\
TIES            & $[0.1,\,1.0]$  & $0.1$ & $20\%$ (fixed) \\
TSV-M           & $[0.1,\,1.0]$  & $0.1$ & -- \\
\bottomrule
\end{tabular}
\caption{\textbf{Hyperparameter tuning for language-model merging.}
We search the merging coefficient $\lambda$, while keeping TIES top-$K$
fixed at $20\%$. ``--'' denotes a parameter that is not applicable.}
\label{tab:language_hyperparameter_search}
\end{table*}%
}

\newcommand{\TabVisionAll}{%
\begin{table*}[t]
\centering
\scriptsize
\setlength{\tabcolsep}{4pt}
\renewcommand{\arraystretch}{1.15}
\begin{tabular}{l|ccc|ccc|ccc}
\toprule
\multirow{2}{*}{\textbf{Merging Method}}
& \multicolumn{3}{c|}{\textbf{ViT-B/32}}
& \multicolumn{3}{c|}{\textbf{ViT-B/16}}
& \multicolumn{3}{c}{\textbf{ViT-L/14}} \\
\cmidrule(lr){2-4}\cmidrule(lr){5-7}\cmidrule(lr){8-10}
& 8 tasks & 14 tasks & 20 tasks
& 8 tasks & 14 tasks & 20 tasks
& 8 tasks & 14 tasks & 20 tasks \\
\midrule

\rowcolor{lightgray}
Individual
& 92.8 & 90.9 & 91.3
& 94.6 & 92.8 & 93.2
& 95.8 & 94.3 & 94.7 \\
\midrule

TA
& 71.1 & 65.4 & 61.3
& 76.3 & 70.8 & 65.9
& 84.8 & 79.6 & 74.0 \\
\rowcolor{green!7}
TA + RI
& \textbf{80.9} & \textbf{69.2} & --
& \textbf{86.6} & \textbf{78.0} & \textbf{71.0}
& \textbf{91.0} & \textbf{84.4} & \textbf{78.9} \\
\midrule

TIES
& 75.0 & 68.2 & 64.2
& 79.9 & 73.8 & 69.2
& 86.8 & 79.4 & 75.5 \\
\rowcolor{green!7}
TIES + RI
& \textbf{83.8} & \textbf{72.2} & \textbf{67.8}
& \textbf{88.7} & \textbf{80.4} & \textbf{75.0}
& \textbf{92.5} & \textbf{84.5} & \textbf{81.1} \\
\midrule

KnOTS-TIES
& 78.1 & 71.8 & 66.7
& 82.4 & 77.4 & 71.3
& 88.4 & 82.6 & 79.1 \\
\rowcolor{green!7}
KnOTS-TIES + RI
& \textbf{84.1} & \textbf{75.9} & \textbf{69.4}
& \textbf{88.9} & \textbf{82.3} & \textbf{76.2}
& \textbf{93.2} & \textbf{87.4} & \textbf{84.2} \\
\midrule

TSV-M
& 85.7 & 80.0 & 76.6
& 89.0 & 84.5 & 80.5
& 92.9 & 89.2 & 87.7 \\
\rowcolor{green!7}
TSV-M + RI
& \textbf{89.2} & \textbf{83.7} & \textbf{82.1}
& \textbf{92.1} & \textbf{88.1} & \textbf{86.0}
& \textbf{94.7} & \textbf{91.7} & \textbf{91.3} \\
\midrule

Iso-CTS
& 86.7 & 81.5 & \textbf{77.7}
& 91.2 & 86.0 & 82.3
& 94.9 & 90.9 & 90.2 \\
\rowcolor{green!7}
Iso-CTS + RI
& \textbf{87.8} & \textbf{82.5} & \textbf{77.7}
& \textbf{92.3} & \textbf{87.7} & \textbf{84.0}
& \textbf{95.2} & \textbf{91.2} & \textbf{90.5} \\
\midrule

WUDI
& 86.9 & 78.3 & 62.4
& 90.1 & 82.9 & 69.6
& 94.0 & 89.9 & 82.8 \\
\rowcolor{green!7}
WUDI + RI
& \textbf{88.3} & \textbf{82.6} & \textbf{75.3}
& \textbf{91.5} & \textbf{86.7} & \textbf{82.2}
& \textbf{94.5} & \textbf{91.1} & \textbf{89.1} \\
\midrule

TATR
& 73.4 & 67.1 & 62.1
& 78.6 & 71.7 & 66.6
& 85.8 & 80.7 & 75.0 \\
\rowcolor{green!7}
TATR + RI
& \textbf{81.7} & \textbf{70.5} & \textbf{64.9}
& \textbf{86.5} & \textbf{78.2} & \textbf{71.5}
& \textbf{91.7} & \textbf{85.3} & \textbf{80.0} \\
\midrule

AdaMerging
& 81.1 & 74.6 & 69.3
& 84.7 & 77.9 & 73.3
& 91.3 & 87.0 & 81.3 \\
\rowcolor{green!7}
AdaMerging + RI
& \textbf{86.8} & \textbf{78.2} & \textbf{70.0}
& \textbf{91.0} & \textbf{82.3} & \textbf{77.9}
& \textbf{93.9} & \textbf{88.6} & \textbf{84.6} \\
\midrule

Surgery
& 85.0 & 83.0 & 83.0
& 86.9 & 84.8 & 85.0
& 91.1 & 89.8 & 88.7 \\
\rowcolor{green!7}
Surgery + RI
& \textbf{87.2} & \textbf{84.5} & \textbf{84.3}
& \textbf{90.0} & \textbf{87.3} & \textbf{86.4}
& \textbf{93.1} & \textbf{91.4} & \textbf{90.2} \\
\midrule

AdaRank
& 87.7 & 78.1 & 70.5
& 90.7 & 81.3 & 76.6
& 93.8 & 86.8 & 82.1 \\
\rowcolor{green!7}
AdaRank + RI
& \textbf{89.8} & \textbf{81.2} & \textbf{73.2}
& \textbf{92.5} & \textbf{83.4} & \textbf{78.1}
& \textbf{95.0} & \textbf{87.6} & \textbf{85.0} \\

\bottomrule
\end{tabular}
\caption{\textbf{RI improves merging across architectures and task settings.}
RI increases accuracy in $88$ of $89$ available paired comparisons;
Iso-CTS on the 20-task ViT-B/32 setting is unchanged at the reported precision.
Entries report mean accuracy (\%), rounded to one decimal place.
RI rows are highlighted in green and individual experts in gray.
\textbf{Bold} identifies the better result within each available pair,
including ties; ``--'' denotes an unavailable result.}
\label{tab:vision_merging_all_architectures}
\end{table*}%
}

\clearpage
\appendix
\setcounter{table}{0}
\renewcommand{\thetable}{A\arabic{table}}
\ifdefined\theHtable
  \renewcommand{\theHtable}{appendix.\arabic{table}}
\fi

\section{Merging Baselines}
\label{app:baselines}

We describe the merging methods evaluated in this work. All experts within
an experiment share a pretrained initialization $\theta_0$, and task $i$
is represented by the parameter update $\tau_i=\theta_i-\theta_0$.
RI adapts these updates before applying the selected merging method.

\paragraph{Parameter aggregation and sparsification.}
\textbf{Weight averaging} provides a reference rule:
$\theta_{\mathrm{avg}}=\theta_0+\frac{1}{N}\sum_{i=1}^{N}\tau_i$.
\textbf{Task Arithmetic (TA)}~\citep{ilharco2023editing} instead uses
$\theta_{\mathrm{TA}}=\theta_0+\lambda\sum_{i=1}^{N}\tau_i$, where we tune
one shared coefficient $\lambda$.
\textbf{TIES}~\citep{yadav2023resolving} retains the largest-magnitude
parameter updates, elects an aggregate sign for each coordinate, and
averages only the retained updates that agree with that sign.
\textbf{TATR}~\citep{sun2025task} restricts merging to parameter coordinates
with low estimated sensitivity to other tasks' losses, reducing the
influence of updates likely to disrupt other tasks.

\paragraph{Spectral merging.}
\textbf{KnOTS-TIES}~\citep{stoica2024model} uses a joint singular value
decomposition (SVD) of layerwise task updates to express them in an aligned
space, applies TIES to the aligned factors, and reconstructs the merged
update.
\textbf{TSV-M}~\citep{gargiulo2025task} retains leading singular components
of each task matrix, concatenates the retained components, and
decorrelates their left and right singular vectors through orthogonal
Procrustes transformations before reconstruction.
\textbf{Iso-CTS}~\citep{marczak2025no} combines dominant directions from the
sum of task matrices with task-specific directions extracted from the
residual updates. It orthogonalizes these directions and assigns a common
singular-value scale to balance common and task-specific contributions.

\paragraph{Optimization and representation correction.}
\textbf{WUDI}~\citep{cheng2025whoever} optimizes each merged layer by
penalizing differences from the expert layers along directions inferred
from their task matrices. These matrices serve as proxies for layer-input
directions, enabling optimization without input examples.
\textbf{AdaMerging}~\citep{yang2024adamerging} learns task-wise or layer-wise
merging coefficients by minimizing predictive entropy on unlabeled inputs. We use the layerwise Adamering in this work.
\textbf{AdaRank}~\citep{lee2026adarank} uses entropy minimization to learn
which singular components of the task updates to retain, together with
merging coefficients. We use the AdaRank-TA variant in this work.
\textbf{Representation Surgery}~\citep{yang2024representation}, abbreviated
as Surgery, learns task-specific correction modules that align the merged
model's representations with those of the corresponding experts. We use the Surgery with Adamerging variant in this work.

\section{Experimental Setup}
\label{app:setup}

\subsection{Vision benchmarks}
\label{App:vision}
\label{ap:task_names}

We evaluate CLIP ViT-B/32, ViT-B/16, and ViT-L/14 on nested suites of
8, 14, and 20 classification tasks. Each task retains its own
classification head, while the shared backbone parameters are merged.
We report the unweighted mean of per-task classification accuracies.
The suites are constructed as follows:
\begin{itemize}
\raggedright
    \item \textbf{8 tasks:} Cars~\citep{kraus2013cars},
    DTD~\citep{cimpoi2014describing}, EuroSAT~\citep{helber2019eurosat},
    GTSRB~\citep{stallkamp2011german}, MNIST~\citep{lecun1998mnist},
    RESISC45~\citep{cheng2017remote}, SUN397~\citep{xiao2016sun},
    and SVHN~\citep{netzer2011reading}.
    \item \textbf{14 tasks:} The eight tasks above, plus
    CIFAR100~\citep{krizhevsky2009cifar}, STL10~\citep{coates2011analysis},
    Flowers102~\citep{nilsback2008automated},
    Oxford-IIIT Pets~\citep{parkhi2012oxfordpets},
    PCAM~\citep{veeling2018rotation}, and FER2013~\citep{goodfellow2013challenges}.
    \item \textbf{20 tasks:} The fourteen tasks above, plus
    EMNIST~\citep{cohen2017emnist}, CIFAR10~\citep{krizhevsky2009cifar},
    Food101~\citep{bossard2014food}, FashionMNIST~\citep{xiao2017fashion},
    RenderedSST2~\citep{socher2013recursive,radford2021learning},
    and KMNIST~\citep{clanuwat2018deep}.
\end{itemize}

\subsection{Language benchmarks}
\label{app:languagebench}

We use domain-specialized experts from MergeBench~\citep{he2026mergebench}
for Llama-3.1-8B-Instruct and Gemma-2-9B-Instruct. We evaluate each domain of the MergeBench across a five-task subset:

\textbf{Instruction following (IFEval).} IFEval~\citep{zhou2023instruction} tests whether a model follows verifiable formatting and content constraints, such as length limits, required keywords, or forbidden punctuation. Each prompt contains one or more instructions that are checked automatically. We report prompt-level accuracy, where a response is correct only if it satisfies every instruction in the prompt.

\textbf{Mathematical reasoning (GSM8K).} GSM8K~\citep{cobbe2021training} consists of grade-school math word problems that require multi-step arithmetic reasoning. We evaluate in a zero-shot chain-of-thought setting and use the flexible scorer, which extracts the final numerical answer from the generated reasoning and compares it with the reference.

\textbf{Multilingual reasoning (ARC-ml).} ARC-ml~\citep{lai2023okapi} translates the AI2 Reasoning Challenge, a set of multiple-choice grade-school science questions, into multiple languages. We report length-normalized accuracy, which selects the answer choice with the highest log-likelihood normalized by its length.

\textbf{Code generation (MBPP+).} MBPP+ extends MBPP~\citep{austin2021program}, a set of entry-level Python programming problems, with additional test cases for more rigorous functional evaluation. We report pass@1, where a problem is solved if the generated program passes all test cases.

\textbf{Safety (XSTest).} XSTest~\citep{rottger2024xstest} measures both appropriate refusal and exaggerated safety behavior. It contains safe prompts that superficially resemble harmful requests, along with contrasting unsafe prompts. We report accuracy, where a response is correct if the model complies with safe prompts and refuses unsafe ones.

Table~\ref{tab:lang_metrics} summarizes the per-example correctness criterion for each benchmark, which we also use to compute consistency in Section~\ref{sec:consistency}.

\begin{table}[h]
\centering
\begin{tabular}{ll}
\toprule
\textbf{Domain} & \textbf{Per-example correctness criterion} \\
\midrule
IFEval & Prompt-level accuracy \\
GSM8K  & Flexible scorer (0-shot CoT) \\
ARC-ml & Normalized accuracy \\
MBPP+  & Pass@1 \\
XSTest & Accuracy \\
\bottomrule
\end{tabular}
\caption{\textbf{Evaluation metrics for the language benchmarks.} Each metric defines per-example correctness for its benchmark.}
\label{tab:lang_metrics}
\end{table}

\subsection{RI adaptation protocol}
\label{app:ri_adaptation}

We follow the test-time adaptation setting and optimize RI using unlabeled
test inputs. Ground-truth labels are not used in the RI loss. We use KL
divergence as the discrepancy metric and set the interference-loss weight
to $1.0$. We compare alternative discrepancy objectives in
Section~\ref{app:discrepancy_metrics}.
Table~\ref{tab:app_ri_settings} summarizes the adaptation settings.
Each expert is adapted independently, with the original expert and shared
pretrained model providing fixed targets. For language models, we apply
the objective to prompt tokens and expert-generated completions. We also
evaluate alternative sources of unlabeled task inputs and unrelated
auxiliary data in Section~\ref{App:additional_analysis}.

\begin{table}[t]
\centering
\small
\setlength{\tabcolsep}{5pt}
\renewcommand{\arraystretch}{1.15}
\begin{tabular}{@{}lcccc@{}}
\toprule
\textbf{Model family} & \textbf{Steps/expert} & \textbf{Batch size}
& \textbf{Learning rate} & \textbf{GPU} \\
\midrule
CLIP ViTs & $2{,}500$ & $32$ & $10^{-6}$ & A40 \\
Llama-3.1-8B-Instruct & $500$ & $32$ & $10^{-5}$ & H200 \\
Gemma-2-9B-Instruct & $500$ & $32$ & $10^{-5}$ & H200 \\
\bottomrule
\end{tabular}
\caption{\textbf{RI adaptation settings.} The step count applies to each
expert independently. All GPUs are NVIDIA models.}
\label{tab:app_ri_settings}
\end{table}

\subsection{Merging hyperparameter selection}
\label{app:merging_hyperparameters}

\paragraph{Vision.}
We select merging hyperparameters on a validation split using the search
grids in Table~\ref{tab:vision_hyperparameter_search}.
For TIES, we tune $\lambda$ first and then the retained fraction of
updates, denoted by top-$K$. The KnOTS-TIES grid also includes top-$K$.
These hyperparameters control parameter aggregation and are distinct from
the RI loss weight.
\TabVisionHyperparameters

\paragraph{Language.}
We reserve $100$ examples for each task drawn from the test split for selecting merging
hyperparameters in the five-domain language setting. This held-out data is kept seperate for tuning and not included in the test partition used for the evaluations. Table~\ref{tab:language_hyperparameter_search} gives the reported search
ranges.

\TabLanguageHyperparameters

\section{Additional Vision Results}
\label{app:vision_results}

\subsection{Results across architectures and task counts}
\label{app:vision_all_results}

Table~\ref{tab:vision_merging_all_architectures} reports results for ten
merging methods across three architectures and three task suites.
RI improves accuracy in $88$ of $89$ available paired comparisons; the
remaining comparison, Iso-CTS on 20-task ViT-B/32, is unchanged at the
reported precision. An RI-adapted configuration achieves the highest
merged-model accuracy in all nine settings. Improvements extend across
spectral methods, adaptive aggregation, and representation correction,
with the largest paired gain of $12.9$ percentage points for WUDI on
20-task ViT-B/32.
\TabVisionAll

\subsection{Choice of discrepancy metric}
\label{app:discrepancy_metrics}

We compare KL divergence, mean squared error (MSE), total variation (TV),
and cross-entropy (CE) as discrepancy objectives for RI in the vision
setting. Table~\ref{tab:ri_discrepancy} reports the resulting merging
accuracies. KL achieves the highest average across the four merging
methods ($84.2\%$), closely followed by MSE ($84.1\%$) and TV ($84.0\%$),
while CE averages $83.2\%$. The preferred metric depends on the merger:
KL performs best for Task Arithmetic and TIES, whereas MSE performs best
for Iso-CTS and TSV-M. These results support KL as a common default,
although the small average differences among KL, MSE, and TV do not
establish a uniformly superior choice.

\begin{table}[t]
    \centering
    \small
    \setlength{\tabcolsep}{7pt}
    \renewcommand{\arraystretch}{1.15}
    \caption{\textbf{KL gives the highest average, while the best metric
    varies across merging methods.}
    Entries report accuracy (\%) after RI adaptation with each discrepancy
    objective. The final row is the unweighted mean across merging methods.
    \textbf{Bold} marks the highest accuracy in each row.}
    \label{tab:ri_discrepancy}
    \begin{tabular}{lcccc}
        \toprule
        \textbf{Merging method} & \textbf{KL} & \textbf{MSE}
        & \textbf{TV} & \textbf{CE} \\
        \midrule
        Task Arithmetic & \textbf{78.8} & 78.3 & 78.6 & 77.5 \\
        TIES            & \textbf{81.5} & 80.9 & 81.3 & 80.3 \\
        Iso-CTS         & 87.9 & \textbf{88.5} & 87.5 & 86.4 \\
        TSV-M           & 88.6 & \textbf{88.8} & 88.7 & 88.4 \\
        \midrule
        Average         & \textbf{84.2} & 84.1 & 84.0 & 83.2 \\
        \bottomrule
    \end{tabular}
\end{table}

\subsection{Prediction drift under cross-task updates}
\label{app:tv_sweep}

We extend the analysis in Section~\ref{sec:background_motivation} to all
eight vision tasks. For each target expert, we increase the contribution
of other-task updates using a coefficient $\alpha\in[0,1]$, comparing
original and RI-adapted task vectors. The setting $\alpha=0$ recovers the
unperturbed expert. We measure prediction drift using total variation
(TV) distance, averaged over the target task's evaluation inputs:
\begin{equation}
    \operatorname{TV}_i^{(v)}(\alpha)
    = \frac{1}{2|\mathcal E_i|}
    \sum_{x\in\mathcal E_i}
    \left\|p_i^{(v)}(x;\alpha)-p_i^{\mathrm{expert}}(x)\right\|_1,
    \label{eq:app_tv_sweep}
\end{equation}
where $v$ denotes the original or RI-adapted condition, $\mathcal E_i$ is
the evaluation set for task $i$, and both distributions are computed
through that task's classification head. TV lies in $[0,1]$, with zero
indicating identical predictive distributions.

Figure~\ref{fig:tv_sweep_all} shows that RI reduces prediction drift and
substantially limits accuracy degradation as other-task contributions
increase. For example, at $\alpha=1$, SVHN accuracy is $94.9\%$ with
RI-adapted updates, compared with $41.0\%$ with the original updates.
Residual degradation remains on some tasks, indicating that RI reduces
but does not eliminate interference in this diagnostic.
\FigTVSweep

\section{Additional Language Analysis}
\label{app:language_results}

\subsection{Consistency with individual experts}
\label{app:consistency}

Aggregate scores can conceal failures on examples that an individual
expert solves correctly. We therefore measure consistency on each
expert's correctly answered examples. Let $\mathcal C_i$ denote this
subset for task $i$, and let $c_i(m,x)$ indicate whether model $m$ answers
example $x$ correctly under that benchmark's evaluation rule. We define
\begin{equation}
    \operatorname{Consistency}_i(m)
    = \frac{100}{|\mathcal C_i|}
      \sum_{x\in\mathcal C_i} c_i(m,x).
    \label{eq:app_consistency}
\end{equation}
We use the same original-expert subset for each baseline and its RI
counterpart, making the comparison a measure of retained correctness.
This metric does not require the generated responses to match verbatim.

Figure~\ref{fig:consistency_all_methods} reports results for TA, TIES,
TSV-M, AdaMerging, and AdaRank on Llama-3.1-8B-Instruct. RI increases
consistency in $20$ of $25$ method--domain comparisons, with one tie and
four decreases. The largest gains occur on instruction following:
AdaMerging improves from $25.7\%$ to $78.7\%$, and AdaRank from $56.3\%$
to $83.2\%$. Improvements are not uniform: TA decreases on multilingual
reasoning, TIES on code generation, and TSV-M on code generation and
safety. These results show that RI generally improves retention of expert
successes, while leaving room to address the remaining task-specific
regressions.
\FigConsistencyAll

\end{document}